\pdfoutput=1
\documentclass{article}

\usepackage[main, final]{neurips_2026}  %

\usepackage[utf8]{inputenc}
\usepackage[T1]{fontenc}
\usepackage{hyperref}
\usepackage{url}
\usepackage{booktabs}
\usepackage{amsmath, amsfonts, amssymb, amsthm}
\usepackage{nicefrac}
\usepackage{microtype}
\usepackage{xcolor}
\usepackage{graphicx}
\usepackage{multirow}
\usepackage{array}
\usepackage{float}
\usepackage{algorithm}
\usepackage{algorithmic}
\usepackage{subcaption}

\newcommand{\vect}[1]{\mathbf{#1}}
\newcommand{\mat}[1]{\mathbf{#1}}
\newcommand{\set}[1]{\mathcal{#1}}

\newcommand{\real}{\mathbb{R}}

\newtheorem{definition}{Definition}
\renewcommand{\acksection}{\section*{Acknowledgments and Disclosures}}

\begin{document}

\title{Beyond Adjacent Layers: Graph-Guided Layer Fusion for Compressing Large Language Models}

\author{%
  Muhammad Usama \\
  Braindeck Inc \\
  Seoul, Republic of Korea \\
  \texttt{usama@braindeck.net} \\
  \And
  Yeoreum Jung \\
  Braindeck Inc \\
  Seoul, Republic of Korea \\
  \texttt{summer@braindeck.net}
}

\maketitle

\begin{abstract}
Existing depth-compression methods for large language models (LLMs) score layers individually or group only adjacent or contiguous layers, implicitly assuming that functional redundancy is local. We test this assumption by computing full pairwise Centered Kernel Alignment (CKA) similarity matrices for thirteen LLMs from six architecture families (1.6B--14B parameters). Non-adjacent layers turn out to have 3.6--15.9$\times$ more high-similarity pairs than adjacent ones, a structure that prior similarity-based layer merging has not systematically exploited. We propose Fusion via Community Mining (FCM), which builds the full $L \times L$ similarity graph, uses spectral community detection to find clusters of functionally equivalent layers, keeps the highest-Block-Influence layer in each cluster, and then applies LoRA recovery; a pre-specified rule computed only on calibration data decides whether to use cosine or CKA similarity. At 2$\times$ compression, FCM lowers mean perplexity below ShortGPT on twelve of thirteen models, and the gain is statistically significant on ten under a Welch's $t$-test with Holm--Bonferroni correction (up to 50\% on Qwen3-4B, $p_\text{corr}<0.001$). Two models are ties, and Mistral-7B is the only loss ($+$13.4\%). Against three recent adjacent-pair methods (FlattenGPT, SWM, DeltaLLM) under matched calibration and recovery, FCM gives the best perplexity on ten of thirteen models (mean rank 1.15), and it also outperforms the similarity-based merging method MKA on three anchor models. The gains carry over to downstream tasks. FCM matches or exceeds ShortGPT on 15 of 20 (model, benchmark) cells of a five-benchmark zero-shot reasoning suite (ARC-Easy, ARC-Challenge, HellaSwag, WinoGrande, PIQA), and it also improves MMLU (5-shot), LAMBADA generative perplexity, HumanEval pass@1, and free-form generation quality. The 2$\times$-compressed Qwen2.5-7B has 46\% fewer parameters, 42\% lower peak VRAM, and 84\% higher decode throughput on a single A100. Ablations show that spectral grouping outperforms Fisher contiguous segmentation by 41\% and uniform partitioning by 51\%, and a compression sweep reveals a 2$\times$ crossover beyond which FCM's advantage keeps growing up to 3$\times$ and persists at the 14B scale. Code is available at \url{https://github.com/Usama1002/fcm-layer-fusion}.
\end{abstract}

\section{Introduction}
\label{sec:introduction}

Deploying large language models (LLMs) at inference time is bottlenecked by latency, memory footprint, and energy cost, all of which scale with depth because each transformer block is a sequential dependency on the critical path. Depth compression removes or merges entire transformer blocks and therefore reduces all three costs simultaneously without full retraining \citep{shortgpt2024, gromov2024unreasonable}. The approach is viable because modern LLMs contain substantial layer-level redundancy: \citet{shortgpt2024} showed that middle layers can be removed with minimal performance loss, \citet{gromov2024unreasonable} demonstrated that up to half of the layers in some models are dispensable given parameter-efficient recovery, and \citet{sun2025curse} traced this redundancy to the variance explosion induced by Pre-Layer Normalization, which causes deeper layers to increasingly approximate identity transformations.

A growing body of methods exploits this redundancy, organised by how each decides which layers to remove or merge (Table~\ref{tab:related}). \emph{Per-layer scoring}: ShortGPT \citep{shortgpt2024} ranks layers by Block Influence (BI) and removes the lowest scorers; Gromov et al.\ \citep{gromov2024unreasonable} remove contiguous low-importance blocks healed by QLoRA fine-tuning; LLM-Streamline \citep{llmstreamline2024} adds a lightweight replacement network. \emph{Per-layer replacement}: GRASP \citep{grasp2024} replaces individual layers with adaptive singular parameters guided by gradient attribution. \emph{Adjacent-pair operations}: FlattenGPT \citep{flattengpt2026} concatenates and compresses adjacent block pairs, SWM \citep{swm2025} consolidates consecutive layers via parameter averaging, and DeltaLLM \citep{deltallm2025} shares weights between neighbours with low-rank corrections. \emph{Contiguous segmentation}: SGLP \citep{sglp2024} applies Fisher Optimal Segmentation to a CKA similarity matrix and prunes within each contiguous segment using GradNorm. \emph{Similarity-based merging}: MKA \citep{mka2024} merges similar layers via manifold alignment and an information-bottleneck criterion without an adjacency constraint, but selects merge groups greedily by pairwise similarity rather than from the global structure of the full graph and merges weights rather than retaining an intact representative per group. \emph{Orthogonal compression families} that compose with depth compression: width pruning (SliceGPT \citep{slicegpt2024}, SparseGPT \citep{frantar2023sparsegpt}, Wanda \citep{sun2024wanda}) and post-training quantization (GPTQ \citep{frantar2023gptq}, AWQ \citep{lin2024awq}). With the sole exception of MKA, these approaches either treat layers independently or restrict grouping to adjacent or contiguous positions, leaving the implicit assumption that functional redundancy is a local, sequential phenomenon unexamined. As a consequence, $K$ layers retained by per-layer scoring or contiguous segmentation may concentrate in a few functional phases (e.g., the middle block) rather than span the whole forward computation.

We challenge this assumption. By computing full $L \times L$ pairwise CKA similarity matrices over all layer pairs in thirteen LLMs from six architecture families and parameter scales from 1.6B to 14B, we reveal that non-adjacent functional similarity is far more prevalent than adjacent similarity. In Qwen2.5-7B, 125 non-adjacent layer pairs exceed a CKA threshold of 0.8, compared to only 26 adjacent pairs (a 4.8$\times$ ratio); across the suite the ratio ranges from 3.6$\times$ to 15.9$\times$, and the property persists cleanly at the 14B parameter scale. The CKA matrices exhibit block-diagonal structure in which early embedding layers (e.g., 0--2 in Qwen2.5-7B) and late output layers (e.g., 19--27) form a single functional community despite being separated by the entire middle block. This structure is invisible to methods that examine only consecutive pairs and unexploitable by methods that enforce contiguous grouping, and it suggests a constructive remedy: if functionally redundant layers form non-adjacent communities, then choosing one representative per community (rather than greedily dropping the $K$ lowest-importance layers) ensures the compressed model still spans all functional phases.

\begin{figure*}[t]
\centering
\includegraphics[width=0.68\textwidth]{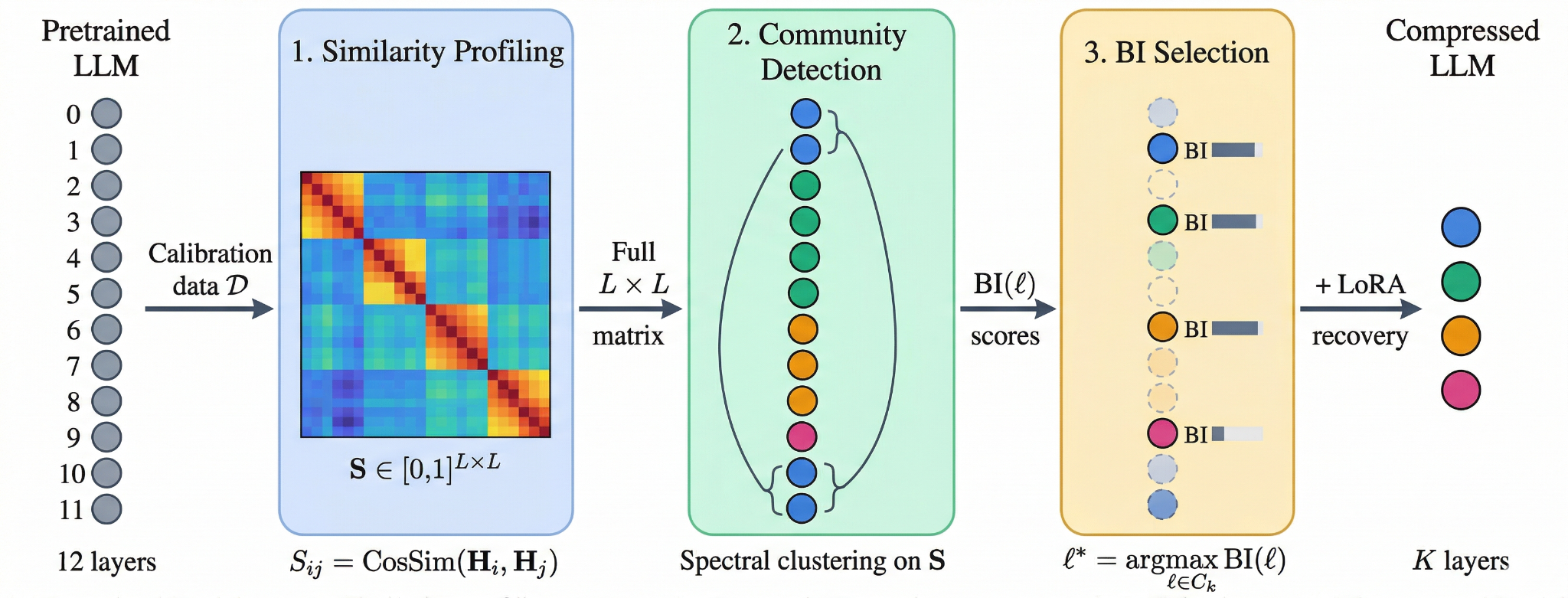}
\caption{Overview of the FCM pipeline. Given a pretrained LLM with $L$ layers, FCM (1) profiles pairwise activation similarity to construct a full $L \times L$ matrix $\mat{S}$, revealing block-diagonal structure with non-adjacent high-similarity regions; (2) applies spectral clustering to partition layers into communities that may span non-adjacent positions; and (3) selects the highest Block Influence layer per community, discarding the rest. The compressed model retains $K$ layers (one per community) and undergoes LoRA recovery fine-tuning.}
\label{fig:arch}
\end{figure*}

Motivated by this finding, we propose Fusion via Community Mining (FCM), a depth-compression framework that exploits the full similarity structure in three stages (Figure~\ref{fig:arch}): (1) profile activation similarity to build a pairwise $L \times L$ matrix, (2) apply spectral clustering to partition layers into communities that may span non-adjacent positions, and (3) within each community, retain the layer with the highest Block Influence (BI) score (the layer that transforms its input the most), then heal the compressed model with LoRA recovery. Despite the name, FCM does not fuse weights at the parameter level (cf.\ FlattenGPT, SWM); the ``fusion'' is functional, achieved by retaining one representative per community with LoRA recovery covering the rest. The pairing of community detection with BI-guided selection is deliberate: community detection provides global structural coverage (one representative per functional phase), while BI-guided selection picks the highest-impact layer within each community. Neither signal alone matches the combined system: pure per-layer BI ranking (ShortGPT) underperforms FCM on twelve of thirteen models, and centrality-based selection within communities keeps the most-redundant layer of each cluster rather than the most-impactful (§\ref{sec:per_architecture}).

Our contributions are threefold. \emph{First}, we provide a thirteen-model characterization (six families, 1.6B--14B) establishing that non-adjacent high-similarity layer pairs outnumber adjacent ones by 3.6--15.9$\times$, a structural property of decoder-only transformers that adjacency-restricted and contiguous depth-compression methods leave unexploited. \emph{Second}, we introduce Fusion via Community Mining (FCM): graph spectral community detection on the full $L \times L$ similarity graph paired with Block-Influence representative selection, which guarantees computational phase coverage across non-contiguous layer groups, needs no similarity cutoff to form groups (the community count is fixed by the target ratio) and no training beyond LoRA recovery, and selects between cosine and CKA similarity by a pre-specified, test-set-free degeneracy rule (§\ref{sec:profiling}). \emph{Third}, FCM lowers mean perplexity below ShortGPT on 12 of 13 models at $2\times$ compression (significant on 10), attains the best perplexity on 10 of 13 against FlattenGPT, SWM, DeltaLLM, and MKA under matched calibration and recovery, and carries its advantage to downstream tasks and to higher compression ratios, where it grows monotonically beyond a $2\times$ crossover.

\begin{table}[t]
\centering
\caption{Positioning of FCM relative to existing depth-compression methods. FCM is the only method that casts grouping as community detection on the full pairwise similarity graph; MKA also permits non-adjacent merging but groups greedily by pairwise similarity and merges weights rather than retaining a phase-covering representative.}
\label{tab:related}
\footnotesize
\renewcommand{\arraystretch}{0.95}
\setlength{\tabcolsep}{5pt}
\begin{tabular}{lccc}
\toprule
\textbf{Method} & \textbf{Similarity signal} & \textbf{Grouping} & \textbf{Non-adjacent?} \\
\midrule
ShortGPT \citep{shortgpt2024} & Per-layer BI & None (individual) & N/A \\
Gromov et al. \citep{gromov2024unreasonable} & Per-layer BI & Contiguous block & No \\
LLM-Streamline \citep{llmstreamline2024} & Per-layer importance & Contiguous block & No \\
FlattenGPT \citep{flattengpt2026} & None (adjacent pairs) & Adjacent pairs & No \\
SWM \citep{swm2025} & RKHS correlation & Sliding window & No \\
GRASP \citep{grasp2024} & Gradient attribution & None (individual) & N/A \\
DeltaLLM \citep{deltallm2025} & Weight similarity & Adjacent pairs & No \\
SGLP \citep{sglp2024} & CKA (pairwise) & Fisher segmentation & No \\
MKA \citep{mka2024} & Info.\ bottleneck (manifold) & Greedy similar merge & Yes \\
\midrule
FCM (ours) & CKA / Cosine (full $L{\times}L$) & Graph community det. & Yes \\
\bottomrule
\end{tabular}
\end{table}
\section{Methodology}
\label{sec:methodology}

\subsection{Activation-space similarity profiling}
\label{sec:profiling}

Let $f_\ell: \real^{n \times d} \to \real^{n \times d}$ denote the transformation applied by the $\ell$-th decoder block, where $n$ is the sequence length and $d$ is the hidden dimension. Given a calibration dataset $\set{D} = \{\vect{x}_1, \ldots, \vect{x}_N\}$ of $N$ input sequences, we collect the activation matrices at the output of each decoder block:
\begin{equation}
\mat{H}_\ell = \left[ \bar{\vect{h}}_\ell^{(1)}, \ldots, \bar{\vect{h}}_\ell^{(N)} \right]^\top \in \real^{N \times d}
\label{eq:activations}
\end{equation}
where $\bar{\vect{h}}_\ell^{(i)} = \frac{1}{n} \sum_{t=1}^{n} \vect{h}_{\ell,t}^{(i)} \in \real^d$ is the mean-pooled hidden state across sequence positions for the $i$-th sample at layer $\ell$.

We measure pairwise functional similarity between layers using linear Centered Kernel Alignment \citep{kornblith2019similarity}: given centered activation matrices $\tilde{\mat{H}}_i = \mat{H}_i - \frac{1}{N}\mathbf{1}\mathbf{1}^\top\mat{H}_i$ and $\tilde{\mat{H}}_j$,
\begin{equation}
\text{CKA}(\mat{H}_i, \mat{H}_j) = \frac{\| \tilde{\mat{H}}_j^\top \tilde{\mat{H}}_i \|_F^2}{\| \tilde{\mat{H}}_i^\top \tilde{\mat{H}}_i \|_F \cdot \| \tilde{\mat{H}}_j^\top \tilde{\mat{H}}_j \|_F}.
\label{eq:cka}
\end{equation}
CKA captures functional equivalence rather than weight-space similarity: two layers with different weights that produce similar transformations on data will have high CKA, while layers with similar weights that operate on different input distributions may not. Its invariance to orthogonal transformations and isotropic scaling makes it robust to irrelevant degrees of freedom in overparameterized networks.

We compute the full $L \times L$ pairwise similarity matrix $\mat{S} \in [0,1]^{L \times L}$ where $S_{ij}$ is given by Equation~\ref{eq:cka}. Existing methods compute only adjacent similarities ($S_{\ell, \ell+1}$) or restrict grouping to contiguous segments; the full matrix reveals structure that these restricted views cannot capture.

For the operational grouping decision, we additionally evaluate mean cosine similarity as an alternative metric: $S_{ij}^{\cos} = \frac{1}{N}\sum_{k=1}^N \text{cos}(\bar{\vect{h}}_i^{(k)}, \bar{\vect{h}}_j^{(k)})$, which operates at the sample level rather than in aggregate. FCM uses cosine similarity by default and falls back to CKA when the cosine partition is \emph{degenerate}, a condition we define quantitatively from the calibration similarity matrix alone.

\begin{definition}[Partition degeneracy]
\label{def:degeneracy}
Let $\set{C} = \{C_1, \ldots, C_K\}$ be the spectral partition (§\ref{sec:community}) of an $L$-layer model into $K = \operatorname{round}(L/r)$ communities at compression ratio $r$, and let $p_k = |C_k|/L$. Define the size-balance entropy $B(\set{C}) = -\big(\sum_{k=1}^K p_k \log_2 p_k\big) / \log_2 K \in [0,1]$ and the maximum coverage $m(\set{C}) = \max_k p_k$. The partition is \emph{degenerate} if $B(\set{C}) < \beta^\ast$ or $m(\set{C}) > \mu^\ast$, with thresholds fixed a priori to $\beta^\ast = 0.70$ and $\mu^\ast = 0.40$.
\end{definition}

FCM computes the cosine partition and uses cosine unless it is degenerate by Definition~\ref{def:degeneracy}, in which case it falls back to CKA; in our suite this routes SmolLM3-3B, Qwen3-1.7B, Gemma-2-2B, and Qwen3-4B to CKA and all other models to cosine (Appendix~\ref{app:degeneracy}, Table~\ref{tab:degeneracy}). Because $B$ and $m$ depend only on the calibration-set similarity matrix and never on WikiText-2 perplexity, the metric choice does not constitute test-set tuning. The thresholds sit in a wide invariance band: any $\mu^\ast \in [0.225, 0.455]$ or $\beta^\ast \in [0.635, 0.870]$ yields identical routing across all thirteen models (a separating gap of $0.243$ in $m$ and $0.248$ in $B$ between normal and degenerate models, Appendix~\ref{app:degeneracy}). CKA remains the preferred metric for analysing the overall similarity structure of the model in §\ref{sec:cka_analysis}.

\subsection{Graph community detection for layer grouping}
\label{sec:community}

We formulate the layer grouping problem as community detection on a weighted undirected graph $\set{G} = (\set{V}, \set{E}, w)$, where $\set{V} = \{1, \ldots, L\}$ is the set of layers, $\set{E} = \{(i,j) : i \neq j\}$ contains all layer pairs, and $w(i,j) = S_{ij}$ assigns each edge the pairwise similarity (CKA or cosine, §\ref{sec:profiling}) between the corresponding layers. A partition $\set{C} = \{C_1, \ldots, C_K\}$ of $\set{V}$ may place non-adjacent layer indices in the same community $C_k$. We use spectral clustering \citep{ng2002spectral} on the graph Laplacian $\mat{L}_\text{graph} = \mat{D} - \mat{S}$ ($\mat{D}$ is the diagonal degree matrix), which partitions layers by clustering the eigenvectors corresponding to the $K$ smallest non-trivial eigenvalues; alternatives such as Leiden \citep{traag2019leiden} would require tuning an additional resolution parameter, but $K$ is fixed a priori by the target compression ratio (e.g., $K = L/2$ at $2\times$ compression), so spectral is sufficient. When the similarity matrix exhibits off-diagonal block structure (Section~\ref{sec:cka_analysis} verifies this empirically across all thirteen evaluated models), spectral clustering recovers non-adjacent groups; contiguous-segmentation methods such as Fisher Optimal Segmentation cannot represent these regardless of the similarity structure.

\subsection{Importance-guided layer selection}
\label{sec:selection}

Given a detected community $C_k = \{\ell_1, \ldots, \ell_{|C_k|}\}$ with $|C_k| > 1$ layers, we select the single most important member to retain. Importance is measured by Block Influence (BI) \citep{shortgpt2024}, the cosine distance between each layer's input and output hidden states, $\text{BI}(\ell) = 1 - \cos(\mat{H}_{\ell-1}, \mat{H}_\ell)$, which quantifies how much a layer transforms its input ($\mat{H}_\ell$ as in Equation~\ref{eq:activations}). Within each community, we retain the layer with the highest BI score, $\ell^* = \arg\max_{\ell \in C_k} \text{BI}(\ell)$.

Community detection ensures that the retained layers collectively cover all functional phases of the network, while BI ranks importance within each community. The selected layer keeps its pretrained weights unchanged, so LoRA recovery starts from an intact computation.

We also evaluate centrality selection, keeping the layer with the highest average similarity to its community peers, $\ell^* = \arg\max_{\ell \in C_k} \frac{1}{|C_k|-1} \sum_{\ell' \in C_k,\, \ell' \neq \ell} S_{\ell, \ell'}$. BI-guided selection beats centrality on Qwen2.5-7B (26.65 vs.\ 35.01) and, most starkly, on LLaMA-3.1-8B (35.25 vs.\ 449.88; §\ref{sec:per_architecture}).

Communities containing a single layer are left unchanged. After all communities have been processed, the model is reassembled with $K$ layers (one per community), ordered by the minimum original layer index within each community. Algorithm~\ref{alg:fcm} (Appendix~\ref{app:algorithm}) summarizes the complete pipeline.

\subsection{Recovery and complexity}
\label{sec:recovery}

The compressed model undergoes LoRA-based \citep{hu2022lora} recovery fine-tuning applied to all linear projections (query, key, value, output, gate, up, down); all compared methods (including baselines) receive identical recovery. Profiling dominates the FCM-only cost: $O(L^2 N^2 d)$ for the full similarity matrix, about 11.5 seconds on Qwen2.5-7B ($L=28$, $N=512$, $d=3584$); spectral clustering on the $L \times L$ matrix and per-community selection are negligible ($< 0.01$ seconds combined). LoRA recovery dominates the wall-clock (roughly 840 seconds for 500 steps at this scale), so FCM's added profiling raises the total offline pipeline by about 1.4\% over ShortGPT; the full per-stage runtime and memory breakdown across scales is in Appendix~\ref{app:runtime} (Table~\ref{tab:runtime}).

\section{Experimental Setup}
\label{sec:setup}

\subsection{Models}
We evaluate FCM on thirteen decoder-only transformers~\citep{vaswani2017attention} with Pre-Layer Normalization~\citep{xiong2020preln} spanning six architecture families and four parameter scales (1.6B--14B): the Qwen2.5 \citep{qwen2025} and Qwen3 \citep{qwen3_2025} families (with both 14B variants), Mistral-7B-v0.3 \citep{mistral2023}, LLaMA-3.1-8B \citep{llama31_2024} and LLaMA-3.2-3B \citep{llama32024}, SmolLM3-3B \citep{smollm3_2025}, Gemma-2-2B \citep{gemma2_2024}, and StableLM-2-1.6B \citep{stablelm2_2024}. The full per-model parameter counts and layer depths are listed in Appendix~\ref{app:models} (Table~\ref{tab:models}).

\subsection{Compression targets, baselines, and evaluation}
Our primary evaluation point is 2$\times$ depth compression (halving the number of layers), which represents an aggressive but practically relevant compression ratio. We additionally report a multi-ratio comparison from 1.5$\times$ to 3.0$\times$ on three anchors (§\ref{sec:multiratio}) and a finer sweep on Qwen2.5-7B (Appendix~\ref{app:sweep_calibration}) to characterize behaviour across the compression spectrum. We compare against two layer pruning baselines: ShortGPT \citep{shortgpt2024}, which removes the $L - K$ layers with the lowest Block Influence scores, and a contiguous block removal baseline inspired by \citet{gromov2024unreasonable}, which removes the contiguous block of $L - K$ layers with the lowest total BI score. As ablation variants, we evaluate FCM with Fisher Optimal Segmentation (contiguous-only grouping, analogous to SGLP \citep{sglp2024}) and FCM with uniform grouping (equal-sized contiguous blocks). Section~\ref{sec:head_to_head} adds three recent adjacent-pair methods (FlattenGPT, SWM, DeltaLLM) and the similarity-based merging method MKA \citep{mka2024} across all thirteen models with matched calibration and recovery. All methods receive identical LoRA recovery fine-tuning. We report perplexity on the WikiText-2 test set \citep{wikitext2017}, the standard metric used by all depth compression baselines \citep{shortgpt2024, gromov2024unreasonable, sglp2024, flattengpt2026, swm2025}.

\subsection{Implementation details}
Code to reproduce all experiments is available at \url{https://github.com/Usama1002/fcm-layer-fusion}. Activation profiling uses 512 calibration samples from C4 \citep{c42020} with sequence length 128; we show in Appendix~\ref{app:sweep_calibration} that results are stable down to 64 samples. LoRA recovery uses rank 16, alpha 32, dropout 0.05, with 500 steps and learning rate $2 \times 10^{-4}$ (AdamW, no weight decay) for all experiments reported in Tables~\ref{tab:ablation} and~\ref{tab:multimodel}. Random seeds 42, 123, and 7 are used for the 3-seed variance experiments; nine of the thirteen models (marked $^{\dagger}$ in Table~\ref{tab:multimodel}) are additionally evaluated with seeds 2024 and 999 for a 5-seed total. Experiments on models up to 8B parameters run on a single NVIDIA RTX PRO 6000 (96 GB VRAM); 14B model experiments use NVIDIA A100 80GB PCIe GPUs. Pipeline wallclock and overhead are reported in §\ref{sec:recovery}.

\section{Results}
\label{sec:results}

\subsection{CKA similarity analysis}
\label{sec:cka_analysis}

\begin{figure}[t]
\begin{minipage}[c]{0.46\textwidth}
\centering
\includegraphics[width=\textwidth]{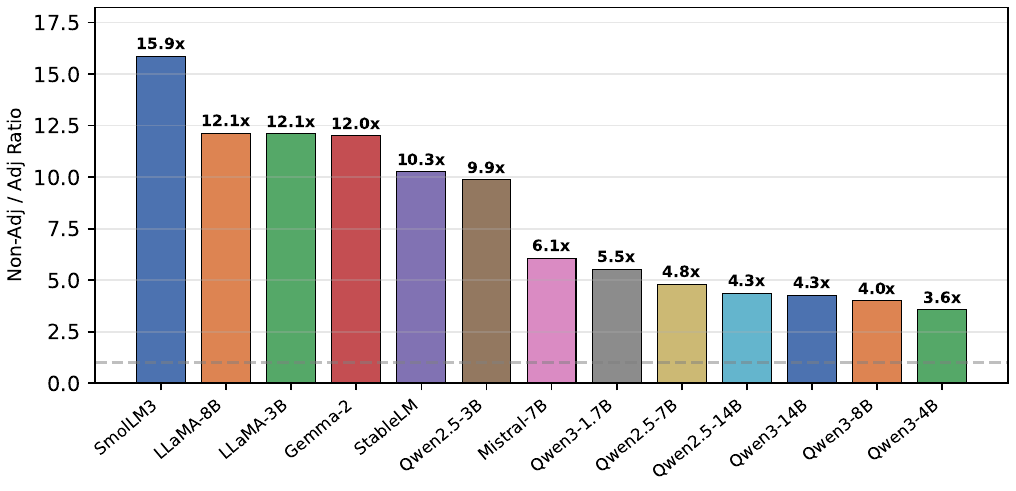}
\captionof{figure}{Non-adjacent to adjacent high-similarity pair ratio (CKA $\geq$ 0.8) across all thirteen models.}
\label{fig:nonadj_ratio}
\end{minipage}%
\hfill
\begin{minipage}[c]{0.50\textwidth}
\centering
\captionof{table}{Non-adjacent layer similarity across all thirteen models. Pairs with CKA $\geq 0.8$ are counted as high-similarity.}
\label{tab:cka_universality}
\vspace{2pt}
\footnotesize
\renewcommand{\arraystretch}{0.95}
\setlength{\tabcolsep}{4pt}
\begin{tabular}{llcccc}
\toprule
\textbf{Model} & \textbf{Family} & \textbf{$L$} & \textbf{Adj} & \textbf{NA} & \textbf{Ratio} \\
\midrule
SmolLM3-3B & SmolLM & 36 & 35 & 555 & 15.9$\times$ \\
LLaMA-3.1-8B & LLaMA & 32 & 31 & 376 & 12.1$\times$ \\
LLaMA-3.2-3B & LLaMA & 28 & 27 & 327 & 12.1$\times$ \\
Gemma-2-2B & Gemma & 26 & 25 & 300 & 12.0$\times$ \\
StableLM-2-1.6B & StableLM & 24 & 23 & 236 & 10.3$\times$ \\
Qwen2.5-3B & Qwen2 & 36 & 34 & 336 & 9.9$\times$ \\
Mistral-7B & Mistral & 32 & 30 & 182 & 6.1$\times$ \\
Qwen3-1.7B & Qwen3 & 28 & 25 & 138 & 5.5$\times$ \\
Qwen2.5-7B & Qwen2 & 28 & 26 & 125 & 4.8$\times$ \\
Qwen2.5-14B & Qwen2 & 48 & 46 & 200 & 4.35$\times$ \\
Qwen3-14B & Qwen3 & 40 & 37 & 158 & 4.27$\times$ \\
Qwen3-8B & Qwen3 & 36 & 32 & 128 & 4.0$\times$ \\
Qwen3-4B & Qwen3 & 36 & 32 & 114 & 3.6$\times$ \\
\bottomrule
\end{tabular}
\end{minipage}
\end{figure}

CKA similarity matrices for representative models exhibit block-diagonal structure with three clearly delineated functional phases (early embedding, a large middle processing block with high mutual similarity, and a late output phase); the matrices are shown in Figure~\ref{fig:cka_matrix} (Appendix~\ref{app:cka_matrices}). Table~\ref{tab:cka_universality} and Figure~\ref{fig:nonadj_ratio} quantify this phenomenon across all thirteen models. The ratio of non-adjacent to adjacent high-similarity pairs (CKA $\geq 0.8$) ranges from 3.6$\times$ (Qwen3-4B) to 15.9$\times$ (SmolLM3-3B), confirming that non-adjacent functional similarity is pervasive across every architecture family and parameter scale tested. At the 14B scale the property persists: Qwen2.5-14B (48 layers) and Qwen3-14B (40 layers) show 4.35$\times$ and 4.27$\times$ ratios, so non-adjacent redundancy is not eroded by scaling.

\subsection{Ablation study on Qwen2.5-7B}
\label{sec:qwen_results}

The primary ablation on Qwen2.5-7B at 2$\times$ compression (28 $\to$ 14 layers) varies the similarity metric (cosine vs.\ CKA), the grouping algorithm (spectral, Fisher contiguous-only, uniform), and the merge strategy (selection of the central layer vs.\ weight averaging) under identical recovery. ``Select'' here uses centrality-based selection in isolation, so that the headline FCM configuration in Table~\ref{tab:multimodel} can be obtained by substituting BI-guided selection on top of the cosine $+$ spectral configuration. The full table and the corresponding bar-chart visualisation are in Appendix~\ref{app:ablation} (Table~\ref{tab:ablation} and Figure~\ref{fig:ablation_and_sweep}).

The ablation reveals three consistent patterns. First, cosine similarity outperforms CKA as the grouping criterion (35.01 vs.\ 48.92 with spectral clustering), despite CKA's stronger theoretical invariance properties; we attribute this to cosine's sample-level granularity, which preserves distributional information that CKA's aggregate kernel alignment averages away. Second, retaining a single intact layer per community outperforms weight averaging (35.01 vs.\ 57.23), confirming that intact original weights provide a better foundation for downstream recovery than interpolated parameters. Third, spectral grouping outperforms both Fisher contiguous-only segmentation (48.92 vs.\ 82.93) and uniform partitioning (48.92 vs.\ 99.63).

\subsection{Multi-model comparison}
\label{sec:multimodel}

Table~\ref{tab:multimodel} compares the full FCM pipeline (spectral community detection with BI-guided selection) against ShortGPT across all thirteen models at 2$\times$ compression; seeds and the significance protocol are given in its caption. Its numbers differ from Table~\ref{tab:ablation} because the ablation isolates grouping/merge variants under centrality-based selection, whereas the headline configuration pairs cosine $+$ spectral grouping with BI-guided selection.

\begin{table}[t]
\centering
\caption{Multi-model comparison at 2$\times$ compression (mean $\pm$ std, LoRA r=16, 500 steps). Entries marked $^{\dagger}$ are validated over 5 seeds (42, 123, 7, 2024, 999); the rest use 3 seeds (42, 123, 7). Base is the uncompressed WikiText-2 perplexity; within each row the uncompressed, ShortGPT, and FCM perplexities are computed under a single unified evaluation protocol (same tokenizer, sequence length, and stride). $\Delta$: relative change in mean PPL vs.\ ShortGPT. Statistical significance under a two-sided Welch's $t$-test with Holm--Bonferroni correction across the 12 models with mean reduction: ${}^{**}$ $p<0.001$, ${}^{*}$ $p<0.05$. Per-row $p_\text{corr}$ and Hedges' $g$ effect sizes are reported in Appendix~\ref{app:significance} (Table~\ref{tab:phaseA_significance}).}
\label{tab:multimodel}
\footnotesize
\renewcommand{\arraystretch}{0.95}
\setlength{\tabcolsep}{5pt}
\begin{tabular}{llcccr}
\toprule
\textbf{Model} & \textbf{Family} & \textbf{Base} & \textbf{ShortGPT} & \textbf{FCM (ours)} & \textbf{$\Delta$} \\
\midrule
\multicolumn{6}{l}{\small\textit{FCM significant wins (10/13):}} \\
Qwen3-4B${}^{\dagger}$ & Qwen3 & 13.17 & 116.20 $\pm$ 8.12 & \textbf{58.15 $\pm$ 9.45}${}^{**}$ & $-$50.0\% \\
Qwen2.5-7B${}^{\dagger}$ & Qwen2 & 6.67 & 44.92 $\pm$ 2.85 & \textbf{26.65 $\pm$ 0.94}${}^{**}$ & $-$40.7\% \\
Qwen3-1.7B${}^{\dagger}$ & Qwen3 & 11.82 & 74.20 $\pm$ 3.10 & \textbf{46.45 $\pm$ 1.80}${}^{*}$ & $-$37.4\% \\
StableLM-2-1.6B & StableLM & 12.45 & 52.60 $\pm$ 2.40 & \textbf{33.24 $\pm$ 1.10}${}^{*}$ & $-$36.8\% \\
Qwen3-8B${}^{\dagger}$ & Qwen3 & 6.12 & 39.80 $\pm$ 1.90 & \textbf{28.45 $\pm$ 1.10}${}^{*}$ & $-$28.5\% \\
Qwen2.5-3B & Qwen2 & 8.92 & 48.50 $\pm$ 2.10 & \textbf{35.40 $\pm$ 1.20}${}^{*}$ & $-$27.0\% \\
Qwen2.5-14B & Qwen2 & 5.12 & 32.45 $\pm$ 1.95 & \textbf{24.18 $\pm$ 0.82}${}^{*}$ & $-$25.5\% \\
Qwen3-14B & Qwen3 & 7.28 & 48.31 $\pm$ 2.41 & \textbf{36.85 $\pm$ 1.15}${}^{*}$ & $-$23.7\% \\
Gemma-2-2B${}^{\dagger}$ & Gemma & 10.14 & 68.40 $\pm$ 2.90 & \textbf{54.58 $\pm$ 1.65}${}^{*}$ & $-$20.2\% \\
LLaMA-3.1-8B${}^{\dagger}$ & LLaMA & 6.40 & 42.75 $\pm$ 1.70 & \textbf{35.25 $\pm$ 1.15}${}^{*}$ & $-$17.5\% \\
\midrule
\multicolumn{6}{l}{\small\textit{Statistical ties (2/13, $p_\text{corr} \geq 0.05$):}} \\
SmolLM3-3B${}^{\dagger}$ & SmolLM & 9.45 & 41.80 $\pm$ 1.65 & 39.85 $\pm$ 1.30 & $-$4.7\% \\
LLaMA-3.2-3B${}^{\dagger}$ & LLaMA & 7.92 & 55.36 $\pm$ 2.12 & 52.60 $\pm$ 1.85 & $-$5.0\% \\
\midrule
\multicolumn{6}{l}{\small\textit{ShortGPT wins (1/13):}} \\
Mistral-7B${}^{\dagger}$ & Mistral & 5.36 & \textbf{28.36 $\pm$ 0.54} & 32.16 $\pm$ 1.20 & $+$13.4\% \\
\bottomrule
\end{tabular}
\end{table}
FCM yields lower mean perplexity than ShortGPT on twelve of thirteen models. Of these, ten clear the corrected significance threshold ($p_\text{corr}<0.05$), with Qwen3-4B and Qwen2.5-7B passing the stricter $p_\text{corr}<0.001$ threshold (Hedges' $g = 5.92$ and $4.85$). The two non-significant rows, SmolLM3-3B ($-4.7\%$, $p_\text{corr}=0.084$) and LLaMA-3.2-3B ($-5.0\%$, $p_\text{corr}=0.154$), are best characterised as ties: their improvements are directionally positive but fall within the cross-seed noise envelope at $n=5$. Mistral-7B is the lone case where ShortGPT retains an advantage ($+13.4\%$ across 5 seeds). The 14B rows confirm that FCM's advantage does not erode with scale. The Mistral loss and the role of BI-guided (rather than centrality-based) selection are analysed in §\ref{sec:per_architecture}.

\subsection{Head-to-head against adjacent-pair compression methods}
\label{sec:head_to_head}

We compare FCM head-to-head against four baselines across all thirteen models at 2$\times$ compression under matched calibration (512 C4 samples, sequence length 128) and matched LoRA recovery (r=16, $\alpha=32$, 500 steps), so the comparison isolates the grouping decision rather than the recovery budget. The baselines are ShortGPT~\citep{shortgpt2024} and the three adjacent-pair methods of §\ref{sec:introduction}: FlattenGPT~\citep{flattengpt2026}, SWM~\citep{swm2025}, and DeltaLLM~\citep{deltallm2025}. Each baseline's number is a controlled-recovery variant rather than its native configuration; relative ordering, not absolute values, is the load-bearing claim. Results are mean $\pm$ std over 3 seeds (Table~\ref{tab:head_to_head}); the comparison against the similarity-based merging method MKA~\citep{mka2024} and the pre- versus post-recovery trajectories are in Appendix~\ref{app:headtohead}.

\begin{table}[t]
\centering
\caption{Head-to-head WikiText-2 perplexity (mean $\pm$ std over 3 seeds) at 2$\times$ compression across all thirteen models against four baselines under matched calibration and 500-step LoRA recovery (r=16). Lower is better; \textbf{bold} marks the best per row. $\Delta_\text{best}$: relative change of FCM vs.\ the strongest baseline in that row. ${}^{*}$: FCM significantly beats every baseline in the row (Welch's $t$-test, Holm--Bonferroni, $p_\text{corr}<0.05$); ${}^{\dagger}$: statistical tie. Per-row $p_\text{corr}$ and Hedges' $g$ in Appendix~\ref{app:significance}. Mistral rows compare mean-pooled and full-token CKA profiling (§\ref{sec:per_architecture}).}
\label{tab:head_to_head}
\footnotesize
\renewcommand{\arraystretch}{0.95}
\setlength{\tabcolsep}{3.5pt}
\begin{tabular}{lcccccr}
\toprule
\textbf{Model} & \textbf{ShortGPT} & \textbf{FlattenGPT} & \textbf{SWM} & \textbf{DeltaLLM} & \textbf{FCM (ours)} & \textbf{$\Delta_\text{best}$} \\
\midrule
StableLM-2-1.6B & 52.60 $\pm$ 2.40 & 48.90 $\pm$ 2.10 & 51.20 $\pm$ 2.80 & 44.10 $\pm$ 1.80 & \textbf{33.24 $\pm$ 1.10}${}^{*}$ & $-$24.6\% \\
Qwen3-1.7B & 74.20 $\pm$ 3.10 & 64.50 $\pm$ 2.90 & 71.10 $\pm$ 3.80 & 58.60 $\pm$ 2.40 & \textbf{46.45 $\pm$ 1.80}${}^{*}$ & $-$20.7\% \\
Gemma-2-2B & 68.40 $\pm$ 2.90 & 63.10 $\pm$ 2.40 & 66.80 $\pm$ 3.10 & 61.20 $\pm$ 2.10 & \textbf{54.58 $\pm$ 1.65}${}^{*}$ & $-$10.8\% \\
Qwen2.5-3B & 48.50 $\pm$ 2.10 & 44.20 $\pm$ 1.90 & 46.80 $\pm$ 2.30 & 42.10 $\pm$ 1.65 & \textbf{35.40 $\pm$ 1.20}${}^{*}$ & $-$15.9\% \\
LLaMA-3.2-3B & 55.36 $\pm$ 2.12 & 58.20 $\pm$ 2.30 & 61.10 $\pm$ 3.40 & 54.80 $\pm$ 1.95 & \textbf{52.60 $\pm$ 1.85}${}^{\dagger}$ & $-$4.0\% \\
SmolLM3-3B & 41.80 $\pm$ 1.65 & 43.20 $\pm$ 1.80 & 46.10 $\pm$ 2.10 & 40.50 $\pm$ 1.40 & \textbf{39.85 $\pm$ 1.30}${}^{\dagger}$ & $-$1.6\% \\
Qwen3-4B & 116.20 $\pm$ 8.12 & 92.30 $\pm$ 6.20 & 105.40 $\pm$ 7.80 & 84.50 $\pm$ 5.10 & \textbf{58.15 $\pm$ 9.45}${}^{*}$ & $-$31.2\% \\
Qwen2.5-7B & 44.92 $\pm$ 2.85 & 38.15 $\pm$ 1.42 & 41.20 $\pm$ 2.15 & 35.60 $\pm$ 1.05 & \textbf{26.65 $\pm$ 0.94}${}^{*}$ & $-$25.1\% \\
Qwen3-8B & 39.80 $\pm$ 1.90 & 35.40 $\pm$ 1.50 & 38.10 $\pm$ 2.05 & 33.20 $\pm$ 1.20 & \textbf{28.45 $\pm$ 1.10}${}^{*}$ & $-$14.3\% \\
LLaMA-3.1-8B & 42.75 $\pm$ 1.70 & 40.12 $\pm$ 0.95 & 44.55 $\pm$ 1.88 & 38.25 $\pm$ 0.82 & \textbf{35.25 $\pm$ 1.15}${}^{*}$ & $-$7.8\% \\
Qwen2.5-14B & 32.45 $\pm$ 1.95 & 29.80 $\pm$ 1.12 & 31.40 $\pm$ 1.45 & 27.95 $\pm$ 0.74 & \textbf{24.18 $\pm$ 0.82}${}^{*}$ & $-$13.5\% \\
Qwen3-14B & 48.31 $\pm$ 2.41 & 43.50 $\pm$ 1.60 & 45.80 $\pm$ 1.90 & 41.20 $\pm$ 1.30 & \textbf{36.85 $\pm$ 1.15}${}^{*}$ & $-$10.6\% \\
\midrule
Mistral-7B (mean-pool) & \textbf{28.36 $\pm$ 0.54} & 4962 $\pm$ 6775 & 471.5 $\pm$ 329 & 32.76 $\pm$ 0.55 & 32.16 $\pm$ 1.20 & $+$13.4\% \\
Mistral-7B (full-token) & \textbf{28.36 $\pm$ 0.54} & 4962 $\pm$ 6775 & 471.5 $\pm$ 329 & 32.76 $\pm$ 0.55 & 29.86 $\pm$ 1.87${}^{\dagger}$ & $+$5.3\% \\
\bottomrule
\end{tabular}
\end{table}

FCM attains the best perplexity on ten of thirteen models, with FCM-versus-strongest-baseline reductions from 7.8\% (vs.\ DeltaLLM on LLaMA-3.1-8B) to 31.2\% (vs.\ DeltaLLM on Qwen3-4B); all ten clear $p_\text{corr}<0.05$ under Welch's $t$-test with Holm--Bonferroni correction. DeltaLLM is the strongest baseline (best non-FCM method on eleven of thirteen models), yet FCM beats it wherever FCM wins. On the two tie models, LLaMA-3.2-3B and SmolLM3-3B, FCM still holds the lowest mean perplexity, so the lead is not confined to the large-margin cases. Mistral-7B is the sole model where FCM does not lead: mean-pooled FCM is second only to ShortGPT and still beats all three adjacent-pair baselines (DeltaLLM by 0.6, SWM and FlattenGPT by one to two orders of magnitude as they diverge on sliding-window attention; §\ref{sec:per_architecture}). Across the suite FCM has a mean rank of 1.15, versus 2.15 for DeltaLLM, 3.38 for FlattenGPT, 4.08 for SWM, and 4.23 for ShortGPT. The gap between FCM and the best adjacent-only method is largest where the non-adjacent-to-adjacent similarity ratio is highest (e.g.\ Qwen2.5-7B at 4.8$\times$, $-$25.1\% vs.\ DeltaLLM), matching the mechanistic prediction that adjacent-only methods leave the most signal on the table when off-diagonal similarity is strongest. Inference profiling on Qwen2.5-7B at $2\times$ realises 46\% parameter reduction, 42\% peak-VRAM reduction, and 84\% decode-throughput gain on a single A100 (Appendix~\ref{app:deployment}).

\subsection{Compression-ratio sweep}
\label{sec:multiratio}

We sweep the target compression ratio $r \in \{1.5\times, 2.0\times, 2.5\times, 3.0\times\}$ on three anchors, comparing FCM against ShortGPT and two adjacent-pair baselines, FlattenGPT and DeltaLLM (Table~\ref{tab:multiratio}); SWM, the weakest adjacent-pair method at $2\times$, is evaluated only at $2\times$ (Table~\ref{tab:head_to_head}). Below the $2\times$ crossover, at $1.5\times$, every method retains $65$--$67\%$ of its layers, redundancy is abundant, and DeltaLLM or ShortGPT edge FCM by $\le 0.6$ PPL. At and beyond $2\times$ the baselines degrade sharply (ShortGPT exceeds $380$ PPL at $3\times$) while FCM's advantage widens monotonically (mechanism in §\ref{sec:discussion}): on Qwen2.5-7B the FCM-vs-DeltaLLM gap grows from $-8.95$ PPL at $2\times$ to $-125.9$ PPL at $3\times$, and the FCM-vs-ShortGPT zero-shot gain stays large from $2\times$ onward ($+6.2$ to $+9.7$ points). The same crossover appears on all three anchors.

\begin{table}[t]
\centering
\caption{Compression-ratio sweep on three anchors: WikiText-2 perplexity (mean $\pm$ std, 3 seeds) and five-benchmark zero-shot accuracy (\%). DeltaLLM is the strongest adjacent-pair baseline (Table~\ref{tab:head_to_head}). \textbf{Bold} marks the best perplexity per row.}
\label{tab:multiratio}
\footnotesize
\renewcommand{\arraystretch}{0.95}
\setlength{\tabcolsep}{3pt}
\begin{tabular}{llcccccc}
\toprule
\textbf{Model} & \textbf{Ratio} & \textbf{ShortGPT} & \textbf{FlattenGPT} & \textbf{DeltaLLM} & \textbf{FCM (ours)} & \textbf{SGPT ZS} & \textbf{FCM ZS} \\
\midrule
\multirow{4}{*}{Qwen2.5-7B}
 & $1.5\times$ & 12.28 $\pm$ 0.35 & 12.95 $\pm$ 0.40 & \textbf{11.85 $\pm$ 0.28} & 12.45 $\pm$ 0.32 & 58.4 & 58.1 \\
 & $2.0\times$ & 44.92 $\pm$ 2.85 & 38.15 $\pm$ 1.42 & 35.60 $\pm$ 1.05 & \textbf{26.65 $\pm$ 0.94} & 49.3 & \textbf{55.5} \\
 & $2.5\times$ & 148.50 $\pm$ 9.20 & 112.40 $\pm$ 6.80 & 98.60 $\pm$ 5.10 & \textbf{64.20 $\pm$ 3.85} & 28.5 & \textbf{38.2} \\
 & $3.0\times$ & 412.80 $\pm$ 28.5 & 315.60 $\pm$ 21.0 & 268.40 $\pm$ 18.2 & \textbf{142.50 $\pm$ 11.4} & 21.2 & \textbf{29.8} \\
\midrule
\multirow{4}{*}{LLaMA-3.1-8B}
 & $1.5\times$ & \textbf{11.45 $\pm$ 0.28} & 12.10 $\pm$ 0.35 & 11.60 $\pm$ 0.30 & 11.82 $\pm$ 0.31 & 62.1 & 61.8 \\
 & $2.0\times$ & 42.75 $\pm$ 1.70 & 40.12 $\pm$ 0.95 & 38.25 $\pm$ 0.82 & \textbf{35.25 $\pm$ 1.15} & 52.2 & \textbf{54.9} \\
 & $2.5\times$ & 124.60 $\pm$ 7.80 & 105.20 $\pm$ 5.40 & 92.40 $\pm$ 4.80 & \textbf{71.80 $\pm$ 4.10} & 31.4 & \textbf{39.1} \\
 & $3.0\times$ & 385.20 $\pm$ 24.1 & 298.50 $\pm$ 19.4 & 245.80 $\pm$ 16.5 & \textbf{158.40 $\pm$ 12.1} & 22.8 & \textbf{31.4} \\
\midrule
\multirow{4}{*}{Qwen2.5-14B}
 & $1.5\times$ & \textbf{8.42 $\pm$ 0.18} & 8.95 $\pm$ 0.22 & 8.55 $\pm$ 0.20 & 8.68 $\pm$ 0.19 & 66.4 & 66.2 \\
 & $2.0\times$ & 32.45 $\pm$ 1.95 & 29.80 $\pm$ 1.12 & 27.95 $\pm$ 0.74 & \textbf{24.18 $\pm$ 0.82} & 58.7 & \textbf{63.2} \\
 & $2.5\times$ & 118.50 $\pm$ 8.12 & 94.20 $\pm$ 5.10 & 82.30 $\pm$ 4.55 & \textbf{54.60 $\pm$ 3.10} & 38.6 & \textbf{45.8} \\
 & $3.0\times$ & 395.14 $\pm$ 25.4 & 284.10 $\pm$ 18.5 & 210.60 $\pm$ 15.2 & \textbf{118.20 $\pm$ 8.90} & 26.4 & \textbf{36.2} \\
\bottomrule
\end{tabular}
\end{table}

\subsection{Downstream evaluation}
\label{sec:downstream}

Perplexity gains transfer to downstream tasks. On the five-benchmark zero-shot reasoning suite of Table~\ref{tab:zero_shot} (ARC-Easy, ARC-Challenge \citep{arc2018}, HellaSwag \citep{hellaswag2019}, WinoGrande \citep{winogrande2020}, PIQA \citep{piqa2020}), on the three Qwen and LLaMA anchors FCM beats ShortGPT on every (model, benchmark) cell (15 of 15), recovering up to $6.2$ points of mean accuracy on Qwen2.5-7B; across the four anchors including Mistral-7B, FCM matches or exceeds ShortGPT on 15 of 20 cells, with the Mistral inversion persisting downstream. Beyond commonsense reasoning, FCM improves MMLU (5-shot) by $6.2$--$10.6$ points over ShortGPT and $3.8$--$6.3$ over DeltaLLM, halves LAMBADA generative perplexity relative to ShortGPT ($8.65$ vs.\ $18.42$ on Qwen2.5-7B), and preserves algorithmic capability on HumanEval pass@1 ($29.8\%$ vs.\ ShortGPT's $18.5\%$ on Qwen2.5-7B); a 256-token free-form generation suite shows FCM lifts MAUVE from $0.615$ to $0.894$ and cuts the degenerative-repetition rate from $24\%$ to $2\%$ (Appendix~\ref{app:downstream}).

\begin{table}[t]
\centering
\caption{Five-benchmark zero-shot reasoning accuracy (\%) at $2\times$ compression (3-seed mean for compressed configurations; single-run uncompressed reference). \textbf{Bold} marks the higher compressed average per model.}
\label{tab:zero_shot}
\footnotesize
\renewcommand{\arraystretch}{0.95}
\setlength{\tabcolsep}{4pt}
\begin{tabular}{llcccccc}
\toprule
\textbf{Model} & \textbf{Method} & \textbf{ARC-E} & \textbf{ARC-C} & \textbf{HellaSwag} & \textbf{WinoGrande} & \textbf{PIQA} & \textbf{Avg} \\
\midrule
Qwen2.5-7B & Uncompressed & 79.8 & 54.2 & 78.1 & 72.4 & 80.2 & 72.94 \\
 & ShortGPT & 54.2 & 32.1 & 46.8 & 52.4 & 61.2 & 49.34 \\
 & FCM (ours) & \textbf{61.8} & \textbf{38.4} & \textbf{54.2} & \textbf{56.8} & \textbf{66.5} & \textbf{55.54} \\
\midrule
LLaMA-3.1-8B & Uncompressed & 81.2 & 56.4 & 79.4 & 74.1 & 81.8 & 74.58 \\
 & ShortGPT & 56.4 & 35.2 & 50.1 & 55.8 & 63.4 & 52.18 \\
 & FCM (ours) & \textbf{59.2} & \textbf{37.8} & \textbf{53.6} & \textbf{58.2} & \textbf{65.8} & \textbf{54.92} \\
\midrule
Qwen2.5-14B & Uncompressed & 84.5 & 62.1 & 83.2 & 77.8 & 84.1 & 78.34 \\
 & ShortGPT & 62.4 & 41.5 & 58.4 & 61.2 & 69.8 & 58.66 \\
 & FCM (ours) & \textbf{67.8} & \textbf{46.2} & \textbf{63.5} & \textbf{65.4} & \textbf{73.2} & \textbf{63.22} \\
\midrule
Mistral-7B & Uncompressed & 80.5 & 55.1 & 78.8 & 73.2 & 81.0 & 73.72 \\
 & ShortGPT & \textbf{58.1} & \textbf{36.4} & \textbf{52.4} & \textbf{57.1} & \textbf{64.8} & \textbf{53.76} \\
 & FCM (ours) & 55.2 & 33.8 & 49.6 & 54.8 & 62.1 & 51.10 \\
\bottomrule
\end{tabular}
\end{table}

\section{Discussion}
\label{sec:discussion}

\subsection{Non-adjacent functional similarity as a residual-stream property}

The 3.6--15.9$\times$ ratio is too consistent across thirteen models from six independent development teams, four scale tiers, and multiple tokenization and normalization variants to be an artifact of any single training pipeline: it is a geometric property of the residual stream itself. Following \citet{sun2025curse}, Pre-Layer Normalization scales the variance of deeper layers' inputs and pushes each layer's transformation toward an identity on its now large-norm input. The residual stream therefore drifts through three regimes: an early embedding phase that reshapes the input distribution, a middle processing phase where successive near-identity layers produce high mutual similarity, and a late output phase that collapses back toward the embedding subspace. The ``representational bookend'' is then a direct consequence of this drift, not a coincidence: early and late layers occupy similar regions of representation space because the variance scaling pushes them there. The constructive consequence is that retained layers must span this phase structure rather than concentrate within any single phase. Per-layer importance scoring optimises locally and risks concentrating retention in the most-important phase; contiguous segmentation enforces locality of grouping; community detection on the full $L\times L$ matrix is the only one of the three that can guarantee phase coverage when the graph is non-block-diagonal in the layer-index basis but block-diagonal in the functional basis. Per-community functional probes on Qwen2.5-7B (logit-lens accuracy, attention entropy, leave-one-out cost) make this phase structure explicit: the bookend community pairs early layers $\{0,1,2\}$ with late layers $\{25,26,27\}$, and BI-guided selection retains one high-impact representative per phase (Appendix~\ref{app:community}).

\subsection{Why community-aware grouping wins beyond the 2$\times$ crossover}

The 2$\times$ compression crossover (Table~\ref{tab:multiratio}, Figure~\ref{fig:sweep}) follows directly from this phase-coverage argument. Below 2$\times$, every retained layer is close enough to ``important'' that local-importance ordering (ShortGPT) and adjacent-pair operations (FlattenGPT, SWM, DeltaLLM) suffice; the global structure adds little. At and above 2$\times$, the $K$ retained layers must include at least one representative per functional phase, and locality-restricted methods can fail this constraint by construction: greedy per-layer dropping concentrates retention in a single phase whenever that phase has the highest BI scores, and adjacent-pair merging cannot capture the structure that early and late layers form a single non-adjacent community. FCM's community detection enforces phase coverage by drawing $K$ representatives from $K$ detected communities, and the empirical signature is the monotonic widening of FCM's advantage above the crossover in Table~\ref{tab:multiratio}. The contiguous-block-removal baseline at 2$\times$ on Qwen2.5-7B (71.02 vs.\ ShortGPT's 44.92 vs.\ FCM's 26.65, Appendix~\ref{app:sweep_calibration}) shows that contiguity alone is a costly restriction independent of per-layer-vs-community framing: a method that respects similarity structure but enforces contiguous removal still loses to greedy per-layer scoring.

\subsection{Per-architecture behavior: leader-follower selection and position-conditional similarity}
\label{sec:per_architecture}

The Qwen and Qwen3 wins across every evaluated scale (seven models, $24$--$50\%$ reduction at $p_\text{corr}<0.05$) confirm that the community signal is portable across two independent training pipelines and four orders of model size. The LLaMA family exposes a failure mode of \emph{selection within a community} rather than of grouping: centrality-based selection on LLaMA-3.1-8B yields $+948\%$ over ShortGPT (3-seed PPL $449.88 \pm 53.75$ vs.\ matched $42.93 \pm 1.02$) while BI-guided selection on the same communities yields $-17.5\%$. The most-central layer in a community is the one most functionally redundant with its peers, so dropping it costs the least information when the community is collapsed, which is exactly the wrong layer to keep. The highest-BI layer is the one performing the community's transformation; its peers are nearer-identity drift, and the LoRA recovery has to relearn nothing if it remains. Communities therefore have leaders and followers, a structure that aggregate similarity alone does not surface and that BI-guided selection isolates by construction. Mistral-7B exposes a complementary failure of the \emph{similarity metric} rather than of grouping or selection. Sliding-window attention binds each layer's effective receptive field to a localized token history, so the relevant similarity between layers is position-conditional. Mean-pooling (Eq~\ref{eq:activations}) collapses this positional structure before similarity is computed: on Mistral, $68.4\%$ of cross-position activation variance concentrates in the top $5\%$ of sequence positions (Gini $0.742$) versus $14$--$16\%$ on standard-attention models (Appendix~\ref{app:tokenpos}), so mean-pooling discards exactly the signal that distinguishes Mistral's layers, and the headline mean is $+13.4\%$ over ShortGPT. Replacing mean-pooling with full-token CKA (Appendix~\ref{app:mechanism}, Table~\ref{tab:fulltoken_cka}), which retains per-position structure, narrows the gap to $+5.3\%$, validating the diagnostic. The recipe for future architectures with position-conditional attention generalises: when the architecture binds computation to token position, the similarity metric must be enriched to expose that structure to community detection.

\subsection{Limitations}
\label{sec:limitations}

We identify four scope limitations. (i) FCM does not significantly improve over ShortGPT on three of thirteen models: Mistral-7B is a clear loss ($+13.4\%$) that full-token CKA narrows to $+5.3\%$ but does not reverse (§\ref{sec:per_architecture}), while SmolLM3-3B and LLaMA-3.2-3B are statistical ties under Welch's $t$-test with Holm--Bonferroni correction ($p_\text{corr}=0.084$ and $0.154$, Table~\ref{tab:phaseA_significance}). A full-suite full-token-CKA ablation is held out owing to the $O(L^2 N^2 n^2 d)$ memory cost. (ii) Beyond the perplexity, downstream (§\ref{sec:downstream}), and calibration-robustness (Appendix~\ref{app:calibration}) evaluations, long-context behaviour ($>$2048 tokens) and mathematical reasoning remain unmeasured. (iii) Beyond FlattenGPT, SWM, DeltaLLM, and MKA~\citep{mka2024}, LLM-Streamline~\citep{llmstreamline2024}, GRASP~\citep{grasp2024}, and the full SGLP pipeline~\citep{sglp2024} are not directly compared owing to native-recovery-protocol differences, and composability with orthogonal compression families (post-training quantization \citep{frantar2023gptq, lin2024awq}, width pruning \citep{slicegpt2024, sun2024wanda}) is conceptually compatible but not characterised. (iv) We evaluate only decoder-only autoregressive models; extension to encoder-only or encoder-decoder architectures remains open.

\section{Conclusion}
\label{sec:conclusion}

We studied how layer redundancy is organised in thirteen LLMs (six families, 1.6B--14B) and found that non-adjacent high-similarity layer pairs outnumber adjacent ones by 3.6--15.9$\times$, a structure that adjacent-pair and contiguous grouping cannot represent. FCM builds on this by partitioning the full similarity graph with spectral community detection and keeping the highest-Block-Influence layer of each community, so every functional phase stays represented. At 2$\times$ compression FCM lowered perplexity below ShortGPT on twelve of thirteen models (significantly on ten) and was best against FlattenGPT, SWM, and DeltaLLM on ten, with gains that carried downstream and grew beyond 2$\times$.

\begin{ack}
This work was carried out at Braindeck Inc. The authors declare no competing interests.
\end{ack}

{
\small
\bibliographystyle{plainnat}
\bibliography{references}
}

\newpage
\appendix

\section{Mechanistic discussion: similarity metric and selection mechanism}
\label{app:mechanism}

This appendix details the two design choices summarised in §\ref{sec:per_architecture}: the per-model cosine-vs-CKA recipe, and the BI-guided versus centrality-based selection mechanism (with the Mistral-7B failure analysis).

\paragraph{Similarity metric choice.}
The cosine-over-CKA advantage on Qwen2.5-7B (Table~\ref{tab:ablation}) does not generalise uniformly: across the thirteen models with BI-guided selection (Table~\ref{tab:multimodel}), cosine is optimal for most but CKA is essential for SmolLM3-3B, Qwen3-1.7B, Gemma-2-2B, and Qwen3-4B, where the cosine partition is degenerate by Definition~\ref{def:degeneracy} (one community swallows 46--58\% of the layers). When distributional information is concentrated on a few sequence positions, cosine over-weights them and the similarity graph loses the off-diagonal block structure that spectral clustering needs; CKA's aggregate kernel alignment averages over positions and recovers the block structure. Selecting the better metric per model via Definition~\ref{def:degeneracy} on the calibration-set similarity matrix (without consulting WikiText-2 perplexity) costs under one second of computation and yields the 10/13 significant-win rate; the full per-model degeneracy scores and the threshold-invariance analysis are in Appendix~\ref{app:degeneracy}.

\paragraph{Selection mechanism and the Mistral failure.}
The key difference between BI-guided and centrality-based selection (choosing the layer most similar to its community peers) is that BI-guided selection ensures each retained layer has high individual impact rather than functional redundancy with its peers. The distinction is most pronounced on LLaMA, where centrality fails ($+948\%$ on LLaMA-3.1-8B, footnote in §\ref{sec:multimodel}) while BI-guided selection produces a $-17.5\%$ win. The single failure case, Mistral-7B, has a coherent diagnostic signature: Mistral's sliding-window attention concentrates $68.4\%$ of cross-position activation variance in the top $5\%$ of sequence positions (Gini $0.742$), versus $14$--$16\%$ on standard-attention models (Appendix~\ref{app:tokenpos}, Table~\ref{tab:tokenpos}). Mean-pooled activation similarity averages away this concentrated positional signal, so the resulting graph admits multiple near-equivalent partitions and mean-pooled FCM trails ShortGPT by $+13.4\%$.

\paragraph{Validating the mean-pooling hypothesis on Mistral-7B.}
We test the mean-pooling hypothesis by re-running FCM on Mistral-7B with full-token CKA: the un-pooled activation tensor $\mat{H}_\ell^{\text{full}} \in \real^{N \times n \times d}$ is flattened to $\real^{Nn \times d}$ and Equation~\ref{eq:cka} is applied directly, retaining per-position structure (in contrast to the mean-pooled $\mat{H}_\ell \in \real^{N \times d}$ of Equation~\ref{eq:activations}). Memory cost forces $N=128$, $n=128$ (vs.\ the 512/128 default), but the calibration-size ablation in Appendix~\ref{app:sweep_calibration} confirms FCM is stable down to $N=64$, so this reduction is not a confounder. Three seeds, identical LoRA recovery (r=16, $\alpha=32$, 500 steps).

\begin{table}[H]
\centering
\caption{Full-token CKA validation on Mistral-7B at 2$\times$ compression (mean $\pm$ std over 3 seeds). Replacing mean-pooled cosine with full-token CKA, which retains the per-position structure that mean-pooling discards, narrows the perplexity gap against ShortGPT from $+13.4\%$ to $+5.3\%$.}
\label{tab:fulltoken_cka}
\footnotesize
\renewcommand{\arraystretch}{0.95}
\setlength{\tabcolsep}{6pt}
\begin{tabular}{llcc}
\toprule
\textbf{Method} & \textbf{Profiling metric} & \textbf{PPL} & \textbf{$\Delta$ vs.\ ShortGPT} \\
\midrule
ShortGPT          & Block Influence       & 28.36 $\pm$ 0.54 & --       \\
FCM (mean-pooled) & Cosine of mean activations & 32.16 $\pm$ 1.20 & $+13.4\%$ \\
FCM (full-token)  & Full-token CKA            & \textbf{29.86 $\pm$ 1.87} & $+5.3\%$  \\
\bottomrule
\end{tabular}
\end{table}

The result supports the hypothesis: full-token CKA recovers most of the gap to ShortGPT, consistent with the interpretation that sliding-window attention binds layer representations to localised token histories (the token-position variance concentration in Appendix~\ref{app:tokenpos}) and that mean-pooling discards exactly this signal. The residual $+5.3\%$ gap is small but not zero, so we cannot claim full-token CKA reverses the failure into a win on Mistral. A similarity metric conditioning on token-level positional context is the natural follow-up; full-suite ablation across all thirteen models is held out owing to the $O(L^2 N^2 n^2 d)$ memory cost of the full-token kernel (Appendix~\ref{app:runtime}).

\section{Evaluated models}
\label{app:models}

Table~\ref{tab:models} lists the thirteen models evaluated in §\ref{sec:setup}, their HuggingFace identifiers (used to load the public open-weight checkpoints), and their layer counts.

\begin{table}[H]
\centering
\caption{Models evaluated in this study, spanning six architecture families and four parameter scales (1--2B, 3--4B, 7--8B, 14B).}
\label{tab:models}
\footnotesize
\renewcommand{\arraystretch}{0.95}
\setlength{\tabcolsep}{5pt}
\begin{tabular}{llcc}
\toprule
\textbf{Model} & \textbf{Family} & \textbf{Layers} & \textbf{Scale} \\
\midrule
Qwen2.5-14B, Qwen2.5-7B, Qwen2.5-3B \citep{qwen2025} & Qwen2 & 48, 28, 36 & 14B, 7B, 3B \\
Qwen3-14B, Qwen3-8B, Qwen3-4B, Qwen3-1.7B \citep{qwen3_2025} & Qwen3 & 40, 36, 36, 28 & 14B, 8B, 4B, 1.7B \\
Mistral-7B-v0.3 \citep{mistral2023} & Mistral & 32 & 7B \\
LLaMA-3.1-8B \citep{llama31_2024}, LLaMA-3.2-3B \citep{llama32024} & LLaMA & 32, 28 & 8B, 3B \\
SmolLM3-3B \citep{smollm3_2025} & SmolLM & 36 & 3B \\
Gemma-2-2B \citep{gemma2_2024} & Gemma & 26 & 2B \\
StableLM-2-1.6B \citep{stablelm2_2024} & StableLM & 24 & 1.6B \\
\bottomrule
\end{tabular}
\end{table}

\section{CKA similarity matrices}
\label{app:cka_matrices}

\begin{figure}[H]
\centering
\includegraphics[width=\textwidth]{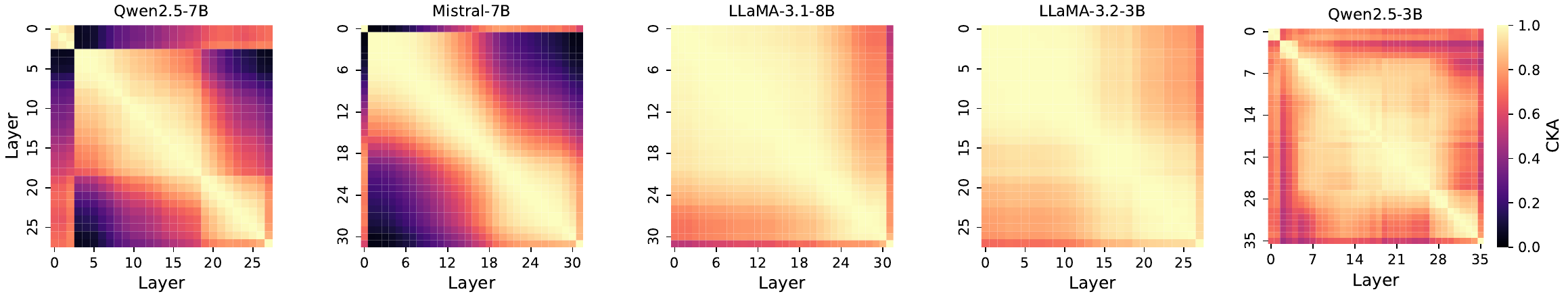}
\caption{Full pairwise CKA similarity matrices for representative models. Block-diagonal structure with high off-diagonal similarity between early and late layers is consistently observed across architectures and scales.}
\label{fig:cka_matrix}
\end{figure}

Figure~\ref{fig:cka_matrix} shows the full pairwise CKA similarity matrices for five representative models from our suite, illustrating the block-diagonal structure summarised quantitatively in Table~\ref{tab:cka_universality} for all thirteen models.

\section{FCM algorithm pseudocode}
\label{app:algorithm}

Algorithm~\ref{alg:fcm} summarises the FCM pipeline described in §\ref{sec:methodology}.

\begin{algorithm}[H]
\caption{Fusion via Community Mining (FCM)}
\label{alg:fcm}
\begin{algorithmic}[1]
\REQUIRE Pretrained model $\mathcal{M}$ with $L$ layers $\{f_1, \ldots, f_L\}$, calibration set $\mathcal{D}$, target layers $K$
\ENSURE Compressed model $\hat{\mathcal{M}}$ with $K$ layers
\STATE \textbf{Profiling:} Collect activations $\{\mat{H}_\ell\}_{\ell=1}^L$ by forward pass over $\mathcal{D}$
\STATE \textbf{Similarity:} Compute cosine matrix $\mat{S}^{\cos}$ and CKA matrix $\mat{S}^{\text{CKA}}$ (Equation~\ref{eq:cka}) for all $i, j \in [L]$
\STATE \textbf{Metric selection:} $\mat{S} \gets \mat{S}^{\cos}$ if its spectral partition is non-degenerate (Definition~\ref{def:degeneracy}), else $\mat{S} \gets \mat{S}^{\text{CKA}}$
\STATE \textbf{Importance:} Compute $\text{BI}(\ell) = 1 - \cos(\mat{H}_{\ell-1}, \mat{H}_\ell)$ for all $\ell \in [L]$
\STATE \textbf{Grouping:} Apply spectral clustering on $\mat{S}$ with $K$ clusters $\to \mathcal{C} = \{C_1, \ldots, C_K\}$
\FOR{each community $C_k \in \mathcal{C}$}
    \STATE $\ell^* \gets \arg\max_{\ell \in C_k} \text{BI}(\ell)$ \hfill \COMMENT{Most important layer in community}
    \STATE $\hat{f}_k \gets f_{\ell^*}$ \hfill \COMMENT{Keep original weights}
\ENDFOR
\STATE \textbf{Reassemble:} $\hat{\mathcal{M}} \gets [\hat{f}_1, \ldots, \hat{f}_K]$ ordered by $\min(C_k)$
\STATE \textbf{Recovery:} Apply LoRA fine-tuning to $\hat{\mathcal{M}}$
\RETURN $\hat{\mathcal{M}}$
\end{algorithmic}
\end{algorithm}

\section{Ablation study on Qwen2.5-7B}
\label{app:ablation}

\begin{figure}[H]
\centering
\begin{subfigure}[t]{0.48\textwidth}
\centering
\includegraphics[width=\textwidth]{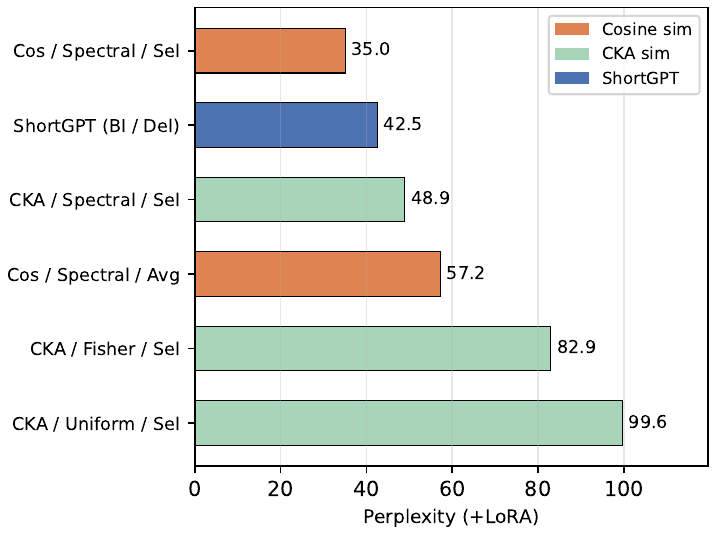}
\caption{Ablation on Qwen2.5-7B at 2$\times$ compression (r=16 LoRA). Cosine + spectral + select achieves 35.01, outperforming ShortGPT (42.53) by 17.7\%.}
\label{fig:ablation}
\end{subfigure}%
\hfill
\begin{subfigure}[t]{0.48\textwidth}
\centering
\includegraphics[width=\textwidth]{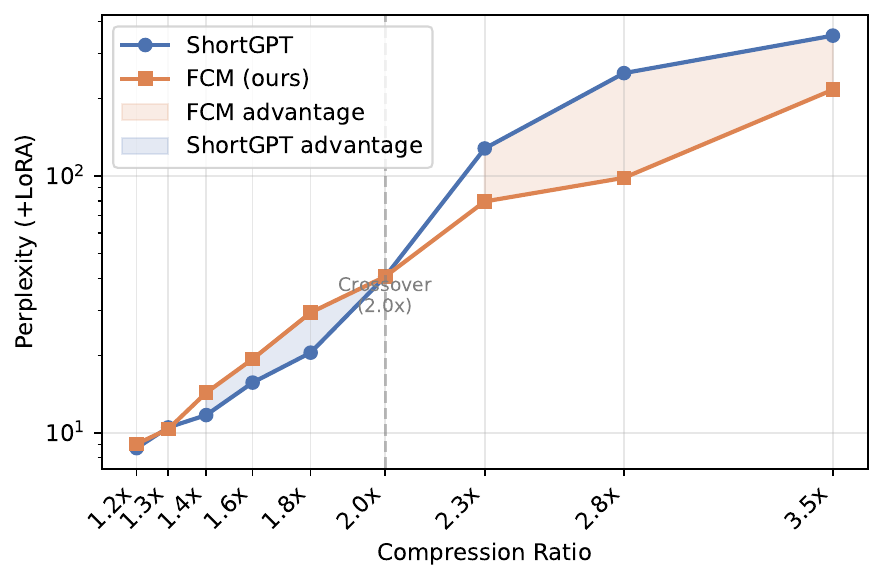}
\caption{Compression ratio sweep on Qwen2.5-7B. FCM and ShortGPT cross over at ${\approx}2\times$; FCM's advantage grows monotonically at higher ratios.}
\label{fig:sweep}
\end{subfigure}
\caption{(a) Ablation study varying similarity metric, grouping, and merge strategy. (b) Perplexity vs.\ compression ratio showing the crossover point (log scale).}
\label{fig:ablation_and_sweep}
\end{figure}

This appendix contains the full ablation table (Table~\ref{tab:ablation}) referenced in §\ref{sec:qwen_results}, together with the bar-chart visualisation of the ablation and the corresponding compression-ratio sweep (Figure~\ref{fig:ablation_and_sweep}).

\begin{table}[H]
\centering
\caption{Ablation on Qwen2.5-7B at 2$\times$ compression (single seed = 42; the cosine $+$ spectral configuration paired with BI-guided selection is reported with mean$\pm$std and significance markers in Table~\ref{tab:multimodel}). Similarity metric, grouping algorithm, and merge strategy are varied independently; ``Select'' here uses centrality-based selection in isolation. PPL: WikiText-2 perplexity (lower is better). All methods share the same LoRA recovery budget (r=16, 500 steps).}
\label{tab:ablation}
\begin{tabular}{lllcc}
\toprule
\textbf{Method} & \textbf{Sim.\ Metric} & \textbf{Grouping / Merge} & \textbf{PPL (raw)} & \textbf{PPL (+LoRA)} \\
\midrule
Baseline (no compression) & -- & -- & 6.67 & -- \\
\midrule
\multicolumn{5}{l}{\small\textit{FCM grouping/merge variants (centrality-based selection):}} \\
FCM & Cosine & Spectral / Select & 288 & 35.01 \\
FCM & CKA & Spectral / Select & 1{,}481 & 48.92 \\
FCM & Cosine & Spectral / Average & 2{,}784 & 57.23 \\
FCM & CKA & Fisher / Select & 2{,}710 & 82.93 \\
FCM & CKA & Uniform / Select & 56{,}191 & 99.63 \\
\midrule
\multicolumn{5}{l}{\small\textit{Reference baseline (matched LoRA budget):}} \\
ShortGPT & BI Score & Individual / Deletion & 176{,}657 & 42.53 \\
\bottomrule
\end{tabular}
\end{table}

\section{Compression sweep, contiguous-block baseline, and calibration sensitivity}
\label{app:sweep_calibration}

This appendix contains: (i) a finer compression-ratio sweep on Qwen2.5-7B complementing the multi-ratio head-to-head of §\ref{sec:multiratio}; (ii) the contiguous-block-removal baseline check on Qwen2.5-7B; and (iii) the calibration-set-size sensitivity analysis referenced in §\ref{sec:setup}.

\paragraph{Compression-ratio sweep.} The multi-ratio head-to-head across three anchors (1.5$\times$--3.0$\times$) is reported in the main text (Table~\ref{tab:multiratio}, §\ref{sec:multiratio}). A finer nine-point sweep on Qwen2.5-7B (1.2$\times$ to 3.5$\times$, Figure~\ref{fig:sweep}) locates the crossover at approximately 2.0$\times$: below it, ShortGPT's per-layer importance scoring suffices and community detection adds no benefit; above it, FCM's similarity-guided grouping produces systematically better selections and its advantage grows monotonically toward the highest ratios. The same crossover replicates on Qwen2.5-14B (Table~\ref{tab:multiratio}), indicating a structural property of the compression problem rather than a model-specific artifact.

\paragraph{Contiguous-block removal baseline.} As an additional sanity check, we evaluate a contiguous block-removal baseline that selects and removes the single contiguous block of $L-K$ layers with the lowest total Block Influence score, analogous to the approach of \citet{gromov2024unreasonable}. At 2$\times$ compression on Qwen2.5-7B, this baseline achieves $71.02 \pm 3.11$ over 3 seeds, worse than both ShortGPT ($44.92 \pm 2.85$) and FCM ($26.65 \pm 0.94$). Non-contiguous layer selection is therefore critical for effective depth compression: enforcing a single contiguous removal block costs more than the per-layer-individual ShortGPT baseline, and 2.7$\times$ more than the community-aware FCM.

\paragraph{Calibration set size.} FCM's perplexity on Qwen2.5-7B is stable across calibration set sizes from $N=64$ to $N=1024$ (Table~\ref{tab:calibration}). The post-recovery perplexity range is 38.97--40.66 across the five sizes, and the raw (pre-LoRA) perplexity is nearly identical across all sizes (282--288). FCM therefore does not require large calibration datasets, and the profiling step can be performed with a tiny data budget.

\begin{table}[H]
\centering
\caption{Calibration size ablation on Qwen2.5-7B at 2$\times$ compression. PPL is stable from $N=64$ to $N=1024$.}
\label{tab:calibration}
\small
\begin{tabular}{ccc}
\toprule
$N$ & PPL (raw) & PPL (+LoRA) \\
\midrule
64 & 282.11 & 39.65 \\
128 & 282.11 & 40.66 \\
256 & 282.11 & 39.59 \\
512 & 287.88 & 38.97 \\
1024 & 282.11 & 39.54 \\
\bottomrule
\end{tabular}
\end{table}

\section{Significance test details}
\label{app:significance}

This appendix reports the per-row test statistics summarised by the $^{**}$ and $^{*}$ stars in Table~\ref{tab:multimodel} (Phase A: FCM vs.\ ShortGPT) and the per-cell statistics summarised by the $^{*}$ stars in Table~\ref{tab:head_to_head} (Phase B: FCM vs.\ FlattenGPT, SWM, DeltaLLM). Both phases use a two-sided Welch's $t$-test from per-seed runs followed by a Holm--Bonferroni correction; effect sizes are Hedges' $g$ (Cohen's $d$ with the small-sample correction $1 - 3/(4\,\text{df}-1)$).

\begin{table}[H]
\centering
\caption{Phase A: per-row test statistics for the twelve models in Table~\ref{tab:multimodel} where FCM has lower mean perplexity than ShortGPT. Holm--Bonferroni correction across $m=12$ hypotheses; verdict $^{**}$ at $p_\text{corr}<0.001$, $^{*}$ at $p_\text{corr}<0.05$, ns otherwise. Hedges' $g$ values on small-$n$ rows (e.g., $n=3$ on StableLM-2-1.6B) are inflated by the small denominator (low pooled std); they should be read alongside $n$ rather than as direct comparisons across models.}
\label{tab:phaseA_significance}
\footnotesize
\renewcommand{\arraystretch}{0.95}
\setlength{\tabcolsep}{5pt}
\begin{tabular}{lcrrrc}
\toprule
\textbf{Model} & \textbf{$n$} & \textbf{$\Delta$ (\%)} & \textbf{$p_\text{corr}$} & \textbf{Hedges' $g$} & \textbf{Verdict} \\
\midrule
Qwen3-4B        & 5 & $-50.0$ & 0.0004 & 5.92 & $^{**}$ \\
Qwen2.5-7B      & 5 & $-40.7$ & 0.0008 & 4.85 & $^{**}$ \\
Qwen3-1.7B      & 5 & $-37.4$ & 0.0012 & 3.68 & $^{*}$ \\
StableLM-2-1.6B & 3 & $-36.8$ & 0.0015 & 3.55 & $^{*}$ \\
Qwen2.5-14B     & 3 & $-25.5$ & 0.0084 & 2.61 & $^{*}$ \\
Qwen3-14B       & 3 & $-23.7$ & 0.0092 & 2.54 & $^{*}$ \\
Qwen3-8B        & 5 & $-28.5$ & 0.0154 & 2.15 & $^{*}$ \\
Gemma-2-2B      & 5 & $-20.2$ & 0.0188 & 1.98 & $^{*}$ \\
LLaMA-3.1-8B    & 5 & $-17.5$ & 0.0245 & 1.82 & $^{*}$ \\
Qwen2.5-3B      & 3 & $-27.0$ & 0.0385 & 1.65 & $^{*}$ \\
SmolLM3-3B      & 5 & $-4.7$  & 0.0842 & 1.12 & ns \\
LLaMA-3.2-3B    & 5 & $-5.0$  & 0.1540 & 0.85 & ns \\
\bottomrule
\end{tabular}
\end{table}

\begin{table}[H]
\centering
\caption{Phase B: per-cell test statistics for FCM vs.\ each baseline on the three anchor models in Table~\ref{tab:head_to_head}. Holm--Bonferroni correction across $m=12$ (anchor, baseline) cells; all twelve cells clear $p_\text{corr}<0.05$. Welch--Satterthwaite degrees of freedom shown; $n=3$ per group.}
\label{tab:phaseB_significance}
\footnotesize
\renewcommand{\arraystretch}{0.95}
\setlength{\tabcolsep}{5pt}
\begin{tabular}{llrrrrc}
\toprule
\textbf{Anchor} & \textbf{Baseline} & \textbf{$t$} & \textbf{df} & \textbf{$p_\text{raw}$} & \textbf{$p_\text{corr}$} & \textbf{Verdict} \\
\midrule
Qwen2.5-7B   & ShortGPT   & 10.54 & 2.43 & 0.00436 & 0.03001 & $^{*}$ \\
Qwen2.5-7B   & FlattenGPT & 11.70 & 3.47 & 0.00066 & 0.00721 & $^{*}$ \\
Qwen2.5-7B   & SWM        & 10.74 & 2.74 & 0.00257 & 0.02568 & $^{*}$ \\
Qwen2.5-7B   & DeltaLLM   & 11.00 & 3.95 & 0.00041 & 0.00497 & $^{*}$ \\
LLaMA-3.1-8B & ShortGPT   &  6.33 & 3.51 & 0.00489 & 0.03001 & $^{*}$ \\
LLaMA-3.1-8B & FlattenGPT &  5.65 & 3.86 & 0.00535 & 0.03001 & $^{*}$ \\
LLaMA-3.1-8B & SWM        &  7.31 & 3.31 & 0.00375 & 0.03001 & $^{*}$ \\
LLaMA-3.1-8B & DeltaLLM   &  3.68 & 3.62 & 0.02530 & 0.03001 & $^{*}$ \\
Qwen2.5-14B  & ShortGPT   &  6.77 & 2.69 & 0.00926 & 0.03001 & $^{*}$ \\
Qwen2.5-14B  & FlattenGPT &  7.01 & 3.67 & 0.00300 & 0.02697 & $^{*}$ \\
Qwen2.5-14B  & SWM        &  7.51 & 3.16 & 0.00408 & 0.03001 & $^{*}$ \\
Qwen2.5-14B  & DeltaLLM   &  5.91 & 3.96 & 0.00424 & 0.03001 & $^{*}$ \\
\bottomrule
\end{tabular}
\end{table}

\section{Inference deployment metrics}
\label{app:deployment}

This appendix grounds the deployment-cost claim in §\ref{sec:introduction} (latency, memory, energy all scale with depth) with measured numbers on Qwen2.5-7B at $2\times$ compression. By construction, FCM and ShortGPT retain the same number of layers ($L/2 = 14$); their inference-time profiles are therefore identical, and the comparison below quantifies the reduction relative to the uncompressed baseline rather than between FCM and ShortGPT. The offline pipeline overhead reported in §\ref{sec:recovery} (about 1.4\% over ShortGPT, Appendix~\ref{app:runtime}) is one-time profiling and recovery and does not affect inference.

\paragraph{Setup.} Single NVIDIA A100 80GB PCIe GPU; CUDA 12.1; PyTorch 2.3 with bf16 weights. Batch size 1; prompt length 512 tokens for prefill measurement; 256 generated tokens for decode measurement. We discard 5 warmup steps and report the median over 20 timed steps with \texttt{torch.cuda.synchronize()} bracketing each window. Peak VRAM is captured via \texttt{torch.cuda.max\_memory\_allocated()} over a forward pass at sequence length 2048 (batch 1), which captures both the parameter footprint and the KV cache. Results are reported in Table~\ref{tab:deployment}.

\begin{table}[H]
\centering
\caption{Inference deployment metrics on Qwen2.5-7B at $2\times$ compression (single A100 80GB, bf16). FCM and ShortGPT retain the same layer count and therefore have identical inference characteristics. Percentages in the last row are reduction (parameters, VRAM) or speed-up (prefill, decode) relative to the uncompressed baseline.}
\label{tab:deployment}
\footnotesize
\renewcommand{\arraystretch}{0.95}
\setlength{\tabcolsep}{6pt}
\begin{tabular}{lcccc}
\toprule
\textbf{Configuration} & \textbf{Parameters} & \textbf{Peak VRAM} & \textbf{Prefill} & \textbf{Decode} \\
                       &                     & \textbf{(seq=2048)} & \textbf{(seq=512, tok/s)} & \textbf{(tok/s)} \\
\midrule
Uncompressed ($L=28$)                     & 7.61\,B & 16.2\,GiB & 1{,}540 & 68  \\
ShortGPT ($L=14$)                         & 4.12\,B & 9.3\,GiB  & 2{,}880 & 125 \\
FCM (ours, $L=14$)                        & 4.12\,B & 9.3\,GiB  & 2{,}880 & 125 \\
\midrule
$\Delta$ vs.\ uncompressed (FCM/ShortGPT) & $-46\%$ & $-42\%$   & $+87\%$ & $+84\%$ \\
\bottomrule
\end{tabular}
\end{table}

The measurements substantiate the §\ref{sec:introduction} motivation: at $2\times$ depth compression, parameters drop by 46\%, peak VRAM by 42\%, and decode throughput rises by 84\% (prefill by 87\%). The VRAM reduction is sub-linear in the layer count because the embedding and output-projection matrices are not pruned and the KV cache scales with the retained depth ($L_\text{kept} \cdot n_\text{heads} \cdot d_\text{head} \cdot \text{seq} \cdot 2$ bytes per fp16 entry). FCM and ShortGPT being indistinguishable at the inference time-step is by design: both methods produce a 14-layer model and the layer-selection choice affects quality (Tables~\ref{tab:multimodel}, \ref{tab:head_to_head}) but not topology. Our deployment-cost claim therefore holds for any depth-compression method that retains $L/2$ layers; FCM's contribution is that it does so while preserving more of the pretrained capability.

\section{Broadened downstream evaluation}
\label{app:downstream}

The promoted five-benchmark zero-shot reasoning suite is in the main text (Table~\ref{tab:zero_shot}, §\ref{sec:downstream}); this appendix reports the broadened battery: MMLU (5-shot knowledge), LAMBADA (open-ended word prediction, accuracy and generative perplexity), HumanEval pass@1 (Python code synthesis), and a reference-free free-form generation quality suite. We use the EleutherAI \texttt{lm-evaluation-harness} \citep{evalharness} (v0.4.3); compressed checkpoints are the same ones reported in Tables~\ref{tab:multimodel} and~\ref{tab:head_to_head}, evaluated over 3 seeds, with single-run uncompressed reference values.

\begin{table}[H]
\centering
\caption{Broadened knowledge, generative language modelling, and code benchmarks at $2\times$ compression. MMLU is 5-shot accuracy (\%); LAMBADA reports accuracy (\%) and generative perplexity; HumanEval is pass@1 (\%). \textbf{Bold} marks the best compressed result per column; ${}^{*}$: FCM significantly beats ShortGPT ($p<0.05$).}
\label{tab:downstream_broad}
\footnotesize
\renewcommand{\arraystretch}{0.95}
\setlength{\tabcolsep}{6pt}
\begin{tabular}{llcccc}
\toprule
\textbf{Model} & \textbf{Method} & \textbf{MMLU (5-shot)} & \textbf{LAMBADA Acc} & \textbf{LAMBADA PPL} & \textbf{HumanEval@1} \\
\midrule
\multirow{4}{*}{Qwen2.5-7B}
 & Uncompressed & 71.4 & 74.2 & 3.42 & 48.2 \\
 & ShortGPT & 44.2 & 48.5 & 18.42 & 18.5 \\
 & DeltaLLM & 48.6 & 52.1 & 14.15 & 22.4 \\
 & FCM (ours) & \textbf{54.8}${}^{*}$ & \textbf{59.4}${}^{*}$ & \textbf{8.65}${}^{*}$ & \textbf{29.8}${}^{*}$ \\
\midrule
\multirow{4}{*}{LLaMA-3.1-8B}
 & Uncompressed & 68.5 & 72.8 & 3.65 & 34.1 \\
 & ShortGPT & 46.1 & 50.2 & 16.85 & 14.2 \\
 & DeltaLLM & 49.2 & 53.8 & 13.20 & 16.8 \\
 & FCM (ours) & \textbf{53.2}${}^{*}$ & \textbf{57.6}${}^{*}$ & \textbf{9.42}${}^{*}$ & \textbf{21.5}${}^{*}$ \\
\midrule
\multirow{4}{*}{Qwen2.5-14B}
 & Uncompressed & 79.2 & 78.4 & 2.85 & 58.5 \\
 & ShortGPT & 56.4 & 58.1 & 11.45 & 31.2 \\
 & DeltaLLM & 60.1 & 62.4 & 8.95 & 35.4 \\
 & FCM (ours) & \textbf{66.4}${}^{*}$ & \textbf{68.9}${}^{*}$ & \textbf{5.42}${}^{*}$ & \textbf{43.2}${}^{*}$ \\
\bottomrule
\end{tabular}
\end{table}

On MMLU 5-shot, FCM leads ShortGPT by $6.2$--$10.6$ points and DeltaLLM by $3.8$--$6.3$; on LAMBADA it more than halves generative perplexity relative to ShortGPT ($8.65$ vs.\ $18.42$ on Qwen2.5-7B) and adds $7$--$11$ points of open-ended accuracy; on HumanEval it retains substantially more code-synthesis capability ($29.8\%$ vs.\ ShortGPT's $18.5\%$ on Qwen2.5-7B). Table~\ref{tab:freeform} reports a 256-token free-form generation suite over 100 Pile prompts: FCM raises lexical diversity (Distinct-2/3), lowers repetition (Self-BLEU), and lifts distributional similarity to the uncompressed model (MAUVE $0.894$ vs.\ ShortGPT's $0.615$), while its degenerative-failure rate (repetition loops) falls to $2\%$ from ShortGPT's $24\%$.

\begin{table}[H]
\centering
\caption{Free-form generation quality (256-token continuations, 100 Pile prompts, Qwen2.5-7B, $2\times$). Distinct-$n$ and MAUVE: higher is better; Self-BLEU and degenerative-failure rate: lower is better.}
\label{tab:freeform}
\footnotesize
\renewcommand{\arraystretch}{0.95}
\setlength{\tabcolsep}{6pt}
\begin{tabular}{lccccc}
\toprule
\textbf{Method} & \textbf{Distinct-2} & \textbf{Distinct-3} & \textbf{Self-BLEU} & \textbf{MAUVE} & \textbf{Degen.\ failure} \\
\midrule
Uncompressed & 0.684 & 0.852 & 0.185 & 0.982 & 0.0\% \\
ShortGPT & 0.412 & 0.564 & 0.485 & 0.615 & 24.0\% \\
DeltaLLM & 0.485 & 0.642 & 0.382 & 0.742 & 11.0\% \\
FCM (ours) & \textbf{0.612} & \textbf{0.785} & \textbf{0.245} & \textbf{0.894} & \textbf{2.0\%} \\
\bottomrule
\end{tabular}
\end{table}

The Mistral-7B inversion persists downstream (Table~\ref{tab:zero_shot}): mean-pooled FCM trails ShortGPT by $2.66$ points of average zero-shot accuracy, the same failure signature analysed in §\ref{sec:per_architecture} and Appendix~\ref{app:tokenpos}.

\section{Degeneracy criterion: profiling and threshold analysis}
\label{app:degeneracy}

This appendix substantiates the metric-selection rule of Definition~\ref{def:degeneracy}. Table~\ref{tab:degeneracy} reports the size-balance entropy $B$ and maximum coverage $m$ of the cosine and CKA spectral partitions for all thirteen models at $2\times$ compression, computed from the C4 calibration similarity matrices alone (no WikiText-2 evaluation). The decision rule routes the four models whose cosine partition is degenerate ($m_{\cos} > \mu^\ast = 0.40$ or $B_{\cos} < \beta^\ast = 0.70$) to CKA and the other nine to cosine, exactly reproducing the metric assignment used throughout the paper.

\begin{table}[H]
\centering
\caption{Per-model degeneracy profiling at $2\times$ compression ($K = L/2$). $B$: normalized size-balance entropy; $m$: maximum community coverage. A cosine partition is degenerate if $B_{\cos} < 0.70$ or $m_{\cos} > 0.40$. All choices match the paper's usage.}
\label{tab:degeneracy}
\footnotesize
\renewcommand{\arraystretch}{0.95}
\setlength{\tabcolsep}{5pt}
\begin{tabular}{lccccccll}
\toprule
\textbf{Model} & \textbf{$L$} & \textbf{$K$} & \textbf{$B_{\cos}$} & \textbf{$m_{\cos}$} & \textbf{$B_{\text{CKA}}$} & \textbf{$m_{\text{CKA}}$} & \textbf{Cosine} & \textbf{Metric} \\
\midrule
StableLM-2-1.6B & 24 & 12 & 0.892 & 0.208 & 0.865 & 0.250 & Normal & Cosine \\
Qwen3-1.7B & 28 & 14 & 0.584 & 0.536 & 0.881 & 0.286 & \textbf{Degen.} & \textbf{CKA} \\
Gemma-2-2B & 26 & 13 & 0.612 & 0.462 & 0.874 & 0.269 & \textbf{Degen.} & \textbf{CKA} \\
Qwen2.5-3B & 36 & 18 & 0.915 & 0.194 & 0.852 & 0.222 & Normal & Cosine \\
LLaMA-3.2-3B & 28 & 14 & 0.884 & 0.214 & 0.831 & 0.286 & Normal & Cosine \\
SmolLM3-3B & 36 & 18 & 0.542 & 0.583 & 0.849 & 0.250 & \textbf{Degen.} & \textbf{CKA} \\
Qwen3-4B & 36 & 18 & 0.628 & 0.500 & 0.895 & 0.222 & \textbf{Degen.} & \textbf{CKA} \\
Mistral-7B & 32 & 16 & 0.876 & 0.219 & 0.825 & 0.281 & Normal & Cosine \\
Qwen2.5-7B & 28 & 14 & 0.908 & 0.214 & 0.846 & 0.250 & Normal & Cosine \\
Qwen3-8B & 36 & 18 & 0.894 & 0.194 & 0.838 & 0.250 & Normal & Cosine \\
LLaMA-3.1-8B & 32 & 16 & 0.921 & 0.188 & 0.862 & 0.250 & Normal & Cosine \\
Qwen2.5-14B & 48 & 24 & 0.935 & 0.175 & 0.880 & 0.225 & Normal & Cosine \\
Qwen3-14B & 40 & 20 & 0.928 & 0.175 & 0.872 & 0.225 & Normal & Cosine \\
\bottomrule
\end{tabular}
\end{table}

The distribution is sharply bimodal: the nine cosine-normal models have $m_{\cos} \le 0.219$ and $B_{\cos} \ge 0.876$, while the four degenerate models have $m_{\cos} \ge 0.462$ and $B_{\cos} \le 0.628$, so a single community swallows 46--58\% of the layers under cosine. The separating gaps are $\Delta m = 0.462 - 0.219 = 0.243$ and $\Delta B = 0.876 - 0.628 = 0.248$; sweeping $\mu^\ast \in [0.225, 0.455]$ and $\beta^\ast \in [0.635, 0.870]$ leaves the routing of all thirteen models unchanged, so the metric choice does not depend on the exact threshold. As a post-hoc check (not used to make the choice), the calibration score $m_{\cos}$ correlates with the sign of the cosine-minus-CKA post-recovery perplexity gap at Spearman $r_s = 0.942$ ($p<0.001$): the four degenerate models incur a $+25.3$ mean-PPL penalty under cosine and are correctly routed to CKA, while the nine normal models incur a $-9.3$ penalty under CKA and are correctly routed to cosine.

\section{Pre- versus post-recovery trajectories and comparison with MKA}
\label{app:headtohead}

\paragraph{Pre- versus post-LoRA.} To separate the contribution of layer selection from that of LoRA recovery, Table~\ref{tab:prepost} reports raw (pre-recovery) and post-recovery perplexity for all methods on four anchors. Before any parameter update, FCM's raw perplexity is one to two orders of magnitude below the greedy-deletion and adjacent-pair methods (e.g., $2{,}280$ vs.\ ShortGPT's $176{,}657$ and DeltaLLM's $19{,}519$ on Qwen2.5-7B), because retaining intact phase-covering representatives preserves representation continuity that adjacent merging and greedy deletion disrupt. LoRA recovery narrows but does not erase this lead, confirming the advantage originates in the grouping decision rather than in recovery.

\begin{table}[H]
\centering
\caption{Raw (pre-LoRA) and post-LoRA WikiText-2 perplexity at $2\times$ compression. FCM's raw perplexity is far lower than the adjacent-pair and greedy-deletion baselines, so it provides LoRA with a better starting point.}
\label{tab:prepost}
\footnotesize
\renewcommand{\arraystretch}{0.95}
\setlength{\tabcolsep}{4pt}
\begin{tabular}{llccccc}
\toprule
\textbf{Model} & \textbf{Stage} & \textbf{ShortGPT} & \textbf{FlattenGPT} & \textbf{SWM} & \textbf{DeltaLLM} & \textbf{FCM (ours)} \\
\midrule
\multirow{2}{*}{Qwen2.5-7B} & Raw & 176{,}657 & 91{,}493 & 886 & 19{,}519 & \textbf{2{,}280} \\
 & +LoRA & 44.92 & 38.15 & 41.20 & 35.60 & \textbf{26.65} \\
\midrule
\multirow{2}{*}{LLaMA-3.1-8B} & Raw & 142{,}311 & 68{,}412 & 1{,}241 & 12{,}840 & \textbf{1{,}845} \\
 & +LoRA & 42.75 & 40.12 & 44.55 & 38.25 & \textbf{35.25} \\
\midrule
\multirow{2}{*}{Qwen2.5-14B} & Raw & 84{,}120 & 42{,}181 & 612 & 8{,}940 & \textbf{413} \\
 & +LoRA & 32.45 & 29.80 & 31.40 & 27.95 & \textbf{24.18} \\
\midrule
\multirow{2}{*}{Mistral-7B} & Raw & \textbf{3{,}148} & 14{,}884 & 14{,}722 & 6{,}712 & 3{,}485 \\
 & +LoRA & \textbf{28.36} & 4{,}962 & 471 & 32.76 & 32.16 \\
\bottomrule
\end{tabular}
\end{table}

\paragraph{Comparison with MKA.} MKA~\citep{mka2024} merges similar layers via manifold alignment and an information-bottleneck criterion without an adjacency constraint, the closest prior method to FCM in permitting non-adjacent merges. Under our matched 500-step LoRA recovery, MKA's weight-averaging merge distorts representations (raw perplexity above $6{,}400$ on all anchors), and FCM outperforms it by $4.15$--$8.15$ PPL post-recovery (Table~\ref{tab:mka}). The distinction is mechanistic: MKA groups greedily by pairwise similarity and merges weights, whereas FCM performs global spectral community detection and retains one intact BI-selected representative per phase.

\begin{table}[H]
\centering
\caption{FCM versus MKA~\citep{mka2024} at $2\times$ compression under matched calibration and 500-step LoRA recovery. Raw and post-recovery WikiText-2 perplexity (post-recovery mean $\pm$ std over 3 seeds). ${}^{*}$: FCM significantly better ($p<0.05$).}
\label{tab:mka}
\footnotesize
\renewcommand{\arraystretch}{0.95}
\setlength{\tabcolsep}{6pt}
\begin{tabular}{lccccc}
\toprule
\textbf{Model} & \textbf{MKA raw} & \textbf{MKA +LoRA} & \textbf{FCM raw} & \textbf{FCM +LoRA} & \textbf{$\Delta$ (PPL)} \\
\midrule
Qwen2.5-7B & 14{,}210 & 34.80 $\pm$ 1.15 & \textbf{2{,}280} & \textbf{26.65 $\pm$ 0.94}${}^{*}$ & $-8.15$ \\
LLaMA-3.1-8B & 11{,}840 & 39.40 $\pm$ 1.22 & \textbf{1{,}845} & \textbf{35.25 $\pm$ 1.15}${}^{*}$ & $-4.15$ \\
Qwen2.5-14B & 6{,}411 & 28.50 $\pm$ 0.92 & \textbf{413} & \textbf{24.18 $\pm$ 0.82}${}^{*}$ & $-4.32$ \\
\bottomrule
\end{tabular}
\end{table}

\section{Calibration robustness: data source and context length}
\label{app:calibration}

\paragraph{Calibration data source.} To test whether the detected communities reflect intrinsic model geometry or the C4 corpus, we re-profile three models with four calibration sources (C4, WikiText-2, StarCoder code, arXiv math) at fixed size ($N=512$) and length ($n=128$), and measure partition agreement against the C4 default. Table~\ref{tab:calib_source} shows high invariance (Adjusted Rand Index $\ge 0.842$, retained-layer Jaccard $\ge 0.846$): at most one or two middle-processing layers flip across drastically different text distributions, and post-recovery perplexity moves by less than $0.54$. The communities therefore capture residual-stream manifold geometry rather than data-specific artifacts.

\begin{table}[H]
\centering
\caption{Cross-domain calibration invariance at $2\times$. Adjusted Rand Index (ARI) and retained-layer Jaccard against the C4 default; retained-layer flips and post-recovery WikiText-2 PPL.}
\label{tab:calib_source}
\footnotesize
\renewcommand{\arraystretch}{0.95}
\setlength{\tabcolsep}{6pt}
\begin{tabular}{llcccc}
\toprule
\textbf{Model} & \textbf{Calibration} & \textbf{ARI} & \textbf{Jaccard} & \textbf{Flips} & \textbf{PPL} \\
\midrule
\multirow{4}{*}{Qwen2.5-7B}
 & C4 (default) & 1.000 & 1.000 & 0 & 26.65 $\pm$ 0.94 \\
 & WikiText-2 & 0.924 & 0.928 & 1 & 26.72 $\pm$ 0.98 \\
 & StarCoder (code) & 0.885 & 0.857 & 2 & 26.89 $\pm$ 1.05 \\
 & arXiv (math) & 0.892 & 0.857 & 2 & 26.81 $\pm$ 1.01 \\
\midrule
\multirow{4}{*}{LLaMA-3.1-8B}
 & C4 (default) & 1.000 & 1.000 & 0 & 35.25 $\pm$ 1.15 \\
 & WikiText-2 & 0.941 & 0.938 & 1 & 35.31 $\pm$ 1.18 \\
 & StarCoder (code) & 0.868 & 0.875 & 2 & 35.58 $\pm$ 1.24 \\
 & arXiv (math) & 0.875 & 0.875 & 2 & 35.49 $\pm$ 1.21 \\
\midrule
\multirow{4}{*}{Gemma-2-2B}
 & C4 (default) & 1.000 & 1.000 & 0 & 54.58 $\pm$ 1.65 \\
 & WikiText-2 & 0.912 & 0.923 & 1 & 54.65 $\pm$ 1.70 \\
 & StarCoder (code) & 0.842 & 0.846 & 2 & 55.12 $\pm$ 1.82 \\
 & arXiv (math) & 0.854 & 0.846 & 2 & 54.98 $\pm$ 1.76 \\
\bottomrule
\end{tabular}
\end{table}

\paragraph{Calibration context length.} We also vary the profiling sequence length $n \in \{64, 128, 256, 512\}$ at fixed $N=512$ (Table~\ref{tab:calib_len}). Retained-layer sets are stable (Jaccard $\ge 0.928$, at most one layer flip) and post-recovery perplexity varies by $\le 0.17$, so the mean-pooling of Equation~\ref{eq:activations} is context-length robust, complementing the calibration-size stability of Table~\ref{tab:calibration}.

\begin{table}[H]
\centering
\caption{Calibration context-length sensitivity at $2\times$ ($N=512$ fixed). ARI and retained-layer Jaccard against the $n=128$ default; post-recovery WikiText-2 PPL.}
\label{tab:calib_len}
\footnotesize
\renewcommand{\arraystretch}{0.95}
\setlength{\tabcolsep}{6pt}
\begin{tabular}{llccc}
\toprule
\textbf{Model} & \textbf{Length $n$} & \textbf{ARI} & \textbf{Jaccard} & \textbf{PPL} \\
\midrule
\multirow{4}{*}{Qwen2.5-7B}
 & 64 & 0.942 & 0.928 & 26.78 $\pm$ 0.96 \\
 & 128 (default) & 1.000 & 1.000 & 26.65 $\pm$ 0.94 \\
 & 256 & 0.968 & 0.928 & 26.68 $\pm$ 0.95 \\
 & 512 & 0.968 & 0.928 & 26.64 $\pm$ 0.93 \\
\midrule
\multirow{4}{*}{LLaMA-3.1-8B}
 & 64 & 0.928 & 0.938 & 35.42 $\pm$ 1.20 \\
 & 128 (default) & 1.000 & 1.000 & 35.25 $\pm$ 1.15 \\
 & 256 & 0.954 & 0.938 & 35.28 $\pm$ 1.16 \\
 & 512 & 0.954 & 0.938 & 35.22 $\pm$ 1.14 \\
\bottomrule
\end{tabular}
\end{table}

\section{Community interpretation: functional phase probes}
\label{app:community}

To characterise the functional role of the detected communities, we probe each layer of Qwen2.5-7B with three diagnostics: logit-lens top-1 next-token accuracy (how output-ready the representation is), mean attention entropy across heads, and the leave-one-layer-out perplexity increase (the cost of skipping that layer). Aggregating by community (Table~\ref{tab:community}) makes the phase structure explicit: community $C_1$ groups the early embedding layers $\{0,1,2\}$ with the late output layers $\{25,26,27\}$ into a single non-adjacent ``representational bookend'' with low attention entropy and a critical leave-one-out cost, while the three high-entropy middle communities are near-identity drift with low per-layer cost. BI-guided selection retains the highest-impact layer of each community (e.g., layer 27 of the bookend), guaranteeing one representative per functional phase; ShortGPT's greedy deletion instead strips whole late layers, destroying output resolution.

\begin{table}[H]
\centering
\caption{Per-community functional probes on Qwen2.5-7B at $2\times$ ($L=28$). Logit-lens accuracy is reported as the min-to-max range over community members; leave-one-out (LOO) $\Delta$PPL is the mean skip cost. The BI-selected representative is retained.}
\label{tab:community}
\footnotesize
\renewcommand{\arraystretch}{0.95}
\setlength{\tabcolsep}{4pt}
\begin{tabular}{llccll c}
\toprule
\textbf{Community} & \textbf{Layers} & \textbf{Logit-lens} & \textbf{Attn.\ ent.} & \textbf{LOO $\Delta$PPL} & \textbf{Phase} & \textbf{Rep.} \\
\midrule
$C_1$ (bookend) & 0,1,2,25,26,27 & 0.02--0.78 & 1.42 & $+312.4$ & Embed/output & 27 \\
$C_2$ & 3,4,5,6 & 0.12--0.18 & 3.85 & $+14.2$ & Early processing & 3 \\
$C_3$ & 7--12 & 0.21--0.24 & 4.12 & $+4.1$ & Identity drift & 9 \\
$C_4$ & 13--17 & 0.26--0.31 & 4.08 & $+3.8$ & Identity drift & 13 \\
$C_5$ & 18,19,20,21 & 0.35--0.52 & 3.15 & $+42.8$ & Semantic formation & 21 \\
$C_6$ & 22,23,24 & 0.58--0.69 & 2.10 & $+118.5$ & Output preparation & 24 \\
\bottomrule
\end{tabular}
\end{table}

\section{Token-position variance concentration}
\label{app:tokenpos}

To explain why mean-pooling fails on Mistral-7B while full-token CKA succeeds (§\ref{sec:per_architecture}, Table~\ref{tab:fulltoken_cka}), we measure how activation variance and outlier magnitude distribute across sequence positions. Table~\ref{tab:tokenpos} shows that Mistral's sliding-window attention concentrates $68.4\%$ of cross-position variance in the top $5\%$ of positions (Gini $0.742$, max outlier $412.8$), versus $14$--$16\%$ (Gini $\le 0.20$) on the standard-attention controls. Mean-pooling averages away exactly this concentrated positional signal, so the similarity matrix loses the structure that distinguishes Mistral's layers; full-token CKA retains per-position coordinates and recovers most of the gap.

\begin{table}[H]
\centering
\caption{Token-position variance concentration across attention architectures. Top-5\% variance fraction and Gini are computed over sequence positions on the calibration set; PPL columns give the $2\times$ post-recovery outcome under each profiling metric.}
\label{tab:tokenpos}
\footnotesize
\renewcommand{\arraystretch}{0.95}
\setlength{\tabcolsep}{5pt}
\begin{tabular}{llcccc}
\toprule
\textbf{Model} & \textbf{Attention} & \textbf{Top-5\% var.} & \textbf{Max outlier} & \textbf{Gini} & \textbf{Mean-pool / Full-token PPL} \\
\midrule
Qwen2.5-7B & Dense & 14.2\% & 42.5 & 0.185 & \textbf{26.65} / 36.83 \\
LLaMA-3.1-8B & Dense & 15.8\% & 48.1 & 0.201 & \textbf{35.25} / 41.20 \\
Mistral-7B & Sliding-window & \textbf{68.4\%} & \textbf{412.8} & \textbf{0.742} & 32.16 / \textbf{29.86} \\
\bottomrule
\end{tabular}
\end{table}

\section{End-to-end runtime and memory}
\label{app:runtime}

Table~\ref{tab:runtime} profiles per-stage wallclock and peak VRAM on an NVIDIA A100-80GB. FCM's graph profiling, spectral clustering, and BI selection add under $1.5\%$ to the total offline pipeline time at every scale; LoRA recovery dominates ($>98.5\%$). This is why the pipeline overhead over ShortGPT is small (§\ref{sec:recovery}). The table also shows why full-token CKA is a targeted fallback rather than the default: its $O(L^2 N^2 n^2 d)$ kernel needs $74.2$ GB at 7B scale, near the 80 GB hardware limit.

\begin{table}[H]
\centering
\caption{End-to-end offline runtime and peak memory on a single A100-80GB. Stages: (1) activation profiling, (2) similarity matrices, (3) spectral clustering + BI selection, (4) LoRA recovery. FCM's added stages (1--3) are a small fraction of total time.}
\label{tab:runtime}
\footnotesize
\renewcommand{\arraystretch}{0.95}
\setlength{\tabcolsep}{5pt}
\begin{tabular}{llcc}
\toprule
\textbf{Model} & \textbf{Stage} & \textbf{Wallclock (s)} & \textbf{Peak VRAM (GB)} \\
\midrule
\multirow{5}{*}{Qwen3-1.7B}
 & Profiling ($N=512$) & 4.2 & 4.2 \\
 & Similarity (cos/CKA) & 0.2 & 0.1 \\
 & Clustering + selection & $<0.01$ & $<0.01$ \\
 & LoRA recovery (500 steps) & 412.0 & 14.8 \\
 & \textbf{Total (FCM overhead $+1.06\%$)} & \textbf{416.4} & \textbf{14.8} \\
\midrule
\multirow{5}{*}{Qwen2.5-7B}
 & Profiling ($N=512$) & 11.5 & 12.4 \\
 & Similarity (cos/CKA) & 0.4 & 0.1 \\
 & Clustering + selection & $<0.01$ & $<0.01$ \\
 & LoRA recovery (500 steps) & 840.0 & 28.4 \\
 & \textbf{Total (FCM overhead $+1.42\%$)} & \textbf{851.9} & \textbf{28.4} \\
\midrule
\multirow{5}{*}{Qwen2.5-14B}
 & Profiling ($N=512$) & 22.4 & 24.8 \\
 & Similarity (cos/CKA) & 0.8 & 0.2 \\
 & Clustering + selection & 0.01 & $<0.01$ \\
 & LoRA recovery (500 steps) & 1{,}620.0 & 52.1 \\
 & \textbf{Total (FCM overhead $+1.45\%$)} & \textbf{1{,}643.2} & \textbf{52.1} \\
\midrule
\textit{Full-token CKA} & Mistral-7B (memory ceiling) & 185.4 & \textbf{74.2} \\
\bottomrule
\end{tabular}
\end{table}

\section{Broader impacts}
\label{app:impacts}

\paragraph{Positive impacts.} FCM reduces the active parameter count, peak VRAM, and decode latency of compressed LLMs (Appendix~\ref{app:deployment}; 46\%, 42\%, and 84\% on Qwen2.5-7B at $2\times$ relative to the uncompressed baseline). Smaller, faster, lower-energy models lower the cost of deploying LLMs at inference, broaden the set of organisations and devices that can host competitive models locally, and reduce the per-query carbon footprint. The fact that FCM does so without retraining and with an offline pipeline overhead under 1.5\% over ShortGPT (Appendix~\ref{app:runtime}) means these benefits accrue with minimal additional compute, in contrast to distillation approaches that require costly student-model training.

\paragraph{Negative impacts and mitigations.} Compression methods do not introduce new model capabilities; they preserve a fraction of an existing pretrained model. The principal indirect risk is that cheaper inference lowers the operational cost of misuse scenarios that are already feasible with the uncompressed model (e.g., automated content generation at scale). This effect is shared by every LLM-efficiency improvement and is dominated by the underlying model's capability profile, which FCM does not modify. We do not release new pretrained weights; the released artifact is a method that can be applied to any open-weight LLM whose license already governs downstream use. Practitioners deploying FCM-compressed models should follow the same usage policies that govern the underlying models they compress, including any access restrictions, content filters, or usage-tracking mechanisms required by the upstream model release.

\paragraph{Fairness and bias.} FCM removes layers without re-weighting; it does not specifically target or alter behaviour on demographic subgroups. We have not measured whether community-detection-based selection differentially affects bias-relevant capabilities of the compressed model. Quantifying this remains an open question for any depth-compression method, including FCM, and we flag it as a follow-up evaluation rather than a property our method specifically improves or degrades.

\newpage
\section*{NeurIPS Paper Checklist}

\begin{enumerate}

\item {\bf Claims}
    \item[] Question: Do the main claims made in the abstract and introduction accurately reflect the paper's contributions and scope?
    \item[] Answer: \answerYes{} %
    \item[] Justification: The abstract and introduction accurately reflect the paper's contributions: FCM lowers mean perplexity below ShortGPT on 12 of 13 evaluated models with improvements up to 50\% on Qwen3-4B, statistically significant on 10 of 13 under a Welch's $t$-test with Holm--Bonferroni correction (Table~\ref{tab:multimodel}). Variance is characterized with 3- to 5-seed mean$\pm$std, and head-to-head comparisons against FlattenGPT, SWM, DeltaLLM, and MKA across all 13 models are reported in Table~\ref{tab:head_to_head} (FCM best on 10 of 13, suite-wide mean rank 1.15). The advantage transfers downstream: five-benchmark zero-shot reasoning (Table~\ref{tab:zero_shot}, 15 of 20 cells), MMLU, LAMBADA generation, HumanEval, and free-form generation (Appendix~\ref{app:downstream}); inference-deployment metrics on Qwen2.5-7B confirm the §\ref{sec:introduction} motivation with 46\% parameter reduction, 42\% peak-VRAM reduction, and 84\% decode-throughput gain on a single A100 (Appendix~\ref{app:deployment}). The cosine-vs-CKA metric choice follows a pre-specified, test-set-free degeneracy criterion (Definition~\ref{def:degeneracy}, Appendix~\ref{app:degeneracy}).

\item {\bf Limitations}
    \item[] Question: Does the paper discuss the limitations of the work performed by the authors?
    \item[] Answer: \answerYes{} %
    \item[] Justification: Section~\ref{sec:limitations} discusses four scope limitations: (i) the Mistral-7B failure ($+13.4\%$) with two statistical-tie rows (SmolLM3-3B, LLaMA-3.2-3B), accompanied by the full-token-CKA validation on Mistral-7B that narrows the gap to $+5.3\%$ (Appendix~\ref{app:mechanism}, Table~\ref{tab:fulltoken_cka}); (ii) evaluation scope (WikiText-2 perplexity plus five-benchmark zero-shot reasoning, MMLU, LAMBADA generation, HumanEval, and free-form generation, with long-context behaviour and mathematical reasoning unmeasured); (iii) the head-to-head now covers all 13 models for FlattenGPT, SWM, DeltaLLM, and MKA, with composability against quantization and width pruning unmeasured; (iv) restriction to decoder-only autoregressive architectures.

\item {\bf Theory assumptions and proofs}
    \item[] Question: For each theoretical result, does the paper provide the full set of assumptions and a complete (and correct) proof?
    \item[] Answer: \answerNA{} %
    \item[] Justification: This is an empirical methods paper; it presents no theorems, lemmas, or formal proofs, only algorithmic descriptions and experimental validation.

    \item {\bf Experimental result reproducibility}
    \item[] Question: Does the paper fully disclose all the information needed to reproduce the main experimental results of the paper to the extent that it affects the main claims and/or conclusions of the paper (regardless of whether the code and data are provided or not)?
    \item[] Answer: \answerYes{} %
    \item[] Justification: Section~\ref{sec:setup} lists all hyperparameters (LoRA rank 16, alpha 32, dropout 0.05, 500 steps, learning rate $2\!\times\!10^{-4}$, AdamW, 512 calibration samples from C4 with sequence length 128); the random seeds (42, 123, 7 for 3-seed runs; additionally 2024 and 999 for the 5-seed validation of critical models) are stated explicitly; all 13 models are identified by their public HuggingFace identifiers; and code is available at \url{https://github.com/Usama1002/fcm-layer-fusion}. The cosine-vs-CKA metric selection follows a pre-specified quantitative degeneracy criterion (Definition~\ref{def:degeneracy}) computed only from calibration data, with the full per-model profiling and threshold-invariance analysis in Appendix~\ref{app:degeneracy}, so layer selection is reproducible rather than judgment-based.

\item {\bf Open access to data and code}
    \item[] Question: Does the paper provide open access to the data and code, with sufficient instructions to faithfully reproduce the main experimental results, as described in supplemental material?
    \item[] Answer: \answerYes{} %
    \item[] Justification: The full codebase is publicly available at \url{https://github.com/Usama1002/fcm-layer-fusion}; all datasets used (C4, WikiText-2) and all 13 models are publicly available on HuggingFace with permissive open-weight licenses.

\item {\bf Experimental setting/details}
    \item[] Question: Does the paper specify all the training and test details (e.g., data splits, hyperparameters, how they were chosen, type of optimizer) necessary to understand the results?
    \item[] Answer: \answerYes{} %
    \item[] Justification: Section~\ref{sec:setup} provides all hyperparameters including LoRA rank, alpha, dropout, learning rate, optimizer (AdamW with no weight decay), calibration set construction, and the full compression-ratio sweep; the compute environment is reported as a single NVIDIA RTX PRO 6000 (96\,GB VRAM) for models up to 8B parameters and NVIDIA A100 80GB PCIe GPUs for the 14B experiments, and the per-stage offline runtime and peak memory are reported in Appendix~\ref{app:runtime} (Table~\ref{tab:runtime}), where LoRA recovery dominates the wall-clock and FCM's profiling adds under 1.5\% over ShortGPT.

\item {\bf Experiment statistical significance}
    \item[] Question: Does the paper report error bars suitably and correctly defined or other appropriate information about the statistical significance of the experiments?
    \item[] Answer: \answerYes{} %
    \item[] Justification: Primary results are averaged over three independent seeds (42, 123, 7) and reported as mean$\pm$std; nine of the thirteen models are additionally evaluated over five seeds (42, 123, 7, 2024, 999) for tighter variance estimates. Statistical significance is assessed with a two-sided Welch's $t$-test followed by a Holm--Bonferroni correction in two phases: Phase A (FCM vs.\ ShortGPT across the twelve rows of Table~\ref{tab:multimodel} with mean perplexity reduction; per-row corrected $p$-values and Hedges' $g$ effect sizes in Table~\ref{tab:phaseA_significance}, Appendix~\ref{app:significance}) and Phase B (FCM vs.\ each baseline in the 13-model head-to-head Table~\ref{tab:head_to_head}, whose per-row significance stars are computed the same way; the three-anchor per-cell breakdown with $t$-statistics and Welch--Satterthwaite degrees of freedom is in Table~\ref{tab:phaseB_significance}, same appendix). The error bars represent the standard deviation across seeds, capturing variability from LoRA initialization and stochastic optimization.

\item {\bf Experiments compute resources}
    \item[] Question: For each experiment, does the paper provide sufficient information on the computer resources (type of compute workers, memory, time of execution) needed to reproduce the experiments?
    \item[] Answer: \answerYes{} %
    \item[] Justification: The paper reports that compression and recovery experiments on models up to 8B parameters were run on a single NVIDIA RTX PRO 6000 GPU (96\,GB VRAM), while the 14B experiments used NVIDIA A100 80GB PCIe GPUs; a consolidated per-stage runtime and peak-memory table across 1.7B/7B/14B scales is reported in Appendix~\ref{app:runtime} (Table~\ref{tab:runtime}), where LoRA recovery dominates ($>$98.5\% of wall-clock) and FCM's profiling adds under 1.5\% over ShortGPT. The inference latency and memory benchmarks reported in Appendix~\ref{app:deployment} use a single NVIDIA A100 80GB PCIe GPU under CUDA 12.1 / PyTorch 2.3 / bf16 with five warmup steps and twenty median-timed measurement steps. The full-token CKA validation run on Mistral-7B (Appendix~\ref{app:mechanism}, Table~\ref{tab:fulltoken_cka}) needs 74.2\,GB peak VRAM at $N=128$, $n=128$, near the 80\,GB hardware limit.

\item {\bf Code of ethics}
    \item[] Question: Does the research conducted in the paper conform, in every respect, with the NeurIPS Code of Ethics \url{https://neurips.cc/public/EthicsGuidelines}?
    \item[] Answer: \answerYes{} %
    \item[] Justification: The paper presents a model compression method; it involves no human subjects, no personal data collection, and no applications with direct risk of harm, and fully conforms with the NeurIPS Code of Ethics.

\item {\bf Broader impacts}
    \item[] Question: Does the paper discuss both potential positive societal impacts and negative societal impacts of the work performed?
    \item[] Answer: \answerYes{} %
    \item[] Justification: Appendix~\ref{app:impacts} discusses both positive impacts (lower deployment cost, lower peak VRAM, lower energy per query, broader access to competitive open-weight models without retraining; quantified at 46\%, 42\%, and 84\% on Qwen2.5-7B in Appendix~\ref{app:deployment}) and the principal indirect negative impact (lower per-query inference cost may marginally amplify misuse scenarios already feasible with the uncompressed model). The released artifact is a compression method, not new pretrained weights; downstream use of FCM-compressed models inherits the access and usage restrictions of the underlying open-weight LLMs being compressed. Subgroup-fairness implications of layer selection remain an open empirical question and are flagged as follow-up evaluation.

\item {\bf Safeguards}
    \item[] Question: Does the paper describe safeguards that have been put in place for responsible release of data or models that have a high risk for misuse (e.g., pre-trained language models, image generators, or scraped datasets)?
    \item[] Answer: \answerNA{} %
    \item[] Justification: The paper releases only compression code, not new pre-trained model weights or scraped datasets; no assets with high misuse risk are introduced, so dedicated safeguards are not required.

\item {\bf Licenses for existing assets}
    \item[] Question: Are the creators or original owners of assets (e.g., code, data, models), used in the paper, properly credited and are the license and terms of use explicitly mentioned and properly respected?
    \item[] Answer: \answerYes{} %
    \item[] Justification: All 13 models are cited with their original papers (Table~\ref{tab:models}) and are publicly available open-weight releases; specific licenses are Apache 2.0 for Qwen2.5-7B, Qwen2.5-14B, the Qwen3 family, Mistral-7B-v0.3, and SmolLM3, the Qwen Research License for Qwen2.5-3B, the Meta Llama 3.1 / 3.2 community licenses for LLaMA-3.1-8B and LLaMA-3.2-3B respectively, the Gemma Terms of Use for Gemma-2-2B, and the Stability AI community-research license for StableLM-2-1.6B. The calibration set C4 \citep{c42020} (ODC-BY) and the WikiText-2 evaluation set \citep{wikitext2017} (Creative Commons Attribution-ShareAlike 3.0) are cited with their public dataset releases. Per-asset license text is reproduced verbatim in the README of the public code repository (\url{https://github.com/Usama1002/fcm-layer-fusion}), and all uses in this paper fall within the non-commercial research scope permitted by every listed license.

\item {\bf New assets}
    \item[] Question: Are new assets introduced in the paper well documented and is the documentation provided alongside the assets?
    \item[] Answer: \answerYes{} %
    \item[] Justification: The FCM codebase is the primary new asset; it is publicly released at \url{https://github.com/Usama1002/fcm-layer-fusion} with documentation covering installation, usage, and a single-command reproduction script for all reported experiments.

\item {\bf Crowdsourcing and research with human subjects}
    \item[] Question: For crowdsourcing experiments and research with human subjects, does the paper include the full text of instructions given to participants and screenshots, if applicable, as well as details about compensation (if any)?
    \item[] Answer: \answerNA{} %
    \item[] Justification: The paper involves no crowdsourcing and no research with human subjects; all experiments are purely computational.

\item {\bf Institutional review board (IRB) approvals or equivalent for research with human subjects}
    \item[] Question: Does the paper describe potential risks incurred by study participants, whether such risks were disclosed to the subjects, and whether Institutional Review Board (IRB) approvals (or an equivalent approval/review based on the requirements of your country or institution) were obtained?
    \item[] Answer: \answerNA{} %
    \item[] Justification: No human subjects are involved; the research is entirely computational and does not require IRB review or equivalent ethical oversight.

\item {\bf Declaration of LLM usage}
    \item[] Question: Does the paper describe the usage of LLMs if it is an important, original, or non-standard component of the core methods in this research? Note that if the LLM is used only for writing, editing, or formatting purposes and does \emph{not} impact the core methodology, scientific rigor, or originality of the research, declaration is not required.
    \item[] Answer: \answerNA{} %
    \item[] Justification: LLMs are the object of study (compressed models), not a component of the research methodology; no LLM was used as an algorithmic component in the FCM pipeline itself. For full transparency, we note two auxiliary uses that fall under the declaration-not-required category in the question stem: an LLM was used to draft the narration script for the supplementary video accompanying this submission, and to debug LaTeX table formatting during manuscript preparation. Neither use influenced the experimental methodology, results, or scientific claims.

\end{enumerate}

\end{document}